\documentclass{article} 
\usepackage[preprint]{colm2026_conference}

\usepackage{microtype}
\usepackage{hyperref}
\usepackage{url}
\usepackage{booktabs}
\usepackage{todonotes}
\usepackage{graphicx}
\usepackage{tabularx}
\usepackage{subcaption}
\usepackage{float}
\usepackage{breakcites}
\usepackage{ifthen}
\usepackage{multirow}
\usepackage[most]{tcolorbox}
\usepackage{enumitem}
\usepackage[switch]{lineno} 
\usepackage{bbm}
\usepackage{xspace} 
\usepackage{amsmath,amsfonts,amssymb}  
\usepackage{booktabs}  
\usepackage{nth}  
\usepackage{csquotes}
\usepackage{cleveref} 
\usepackage{newtxtext,newtxmath}
\usepackage{csquotes}
\usepackage{setspace}
\usepackage{algorithm}
\usepackage{algpseudocode}
\usepackage{mathtools}
\usepackage{listings}
\usepackage{xcolor}
\definecolor{backcolour}{rgb}{0.99,0.99,0.99}
\definecolor{codegray}{rgb}{0.5,0.5,0.5}
\lstdefinestyle{mystyle}{
    backgroundcolor=\color{backcolour},   
    numberstyle=\tiny\color{codegray},
    basicstyle=\ttfamily\footnotesize,
    postbreak=\mbox{\textcolor{gray}{$\hookrightarrow$}\space},
    breakatwhitespace=false,         
    breaklines=true,                 
    captionpos=b,                    
    keepspaces=true,                 
    numbers=left,                    
    numbersep=5pt,                  
    showspaces=false,                
    showstringspaces=false,
    showtabs=false,                  
    tabsize=2,
    frame=single,
    framesep=3pt
}
\newcommand{\LR}{Lyria\xspace}
\newcommand{\GF}{LAFT\xspace}

\definecolor{red}{RGB}{220, 50, 47}
\definecolor{promptbg}{RGB}{240, 248, 255} 
\definecolor{feedbackbg}{RGB}{255, 204, 204}

\newcounter{promptblockcounter}
\renewcommand{\thepromptblockcounter}{\arabic{promptblockcounter}}
\newenvironment{promptblock}[1]
{
    \refstepcounter{promptblockcounter}
    \begin{tcolorbox}[
        enhanced,
        breakable,
        colback=promptbg,
        colframe=blue!55!black,
        boxrule=0.75pt, 
        arc=3pt, 
        title={Prompt Template~\thepromptblockcounter: #1}, 
        before upper=\raggedright,
    ]
}
{
    \end{tcolorbox}
}

\newcounter{feedbackblockcounter}
\renewcommand{\thefeedbackblockcounter}{\arabic{feedbackblockcounter}}

\usepackage{lineno}

\definecolor{darkblue}{rgb}{0, 0, 0.5}
\hypersetup{colorlinks=true, citecolor=darkblue, linkcolor=darkblue, urlcolor=darkblue}

\title{Lyria: A Genetic Algorithm-Driven Neuro-Symbolic Reasoning Framework for LLMs}

\author{Weizhi Tang, Kwabena Nuamah, Vaishak Belle \\
Artificial Intelligence Applications Institute\\
The University of Edinburgh\\
\texttt{\{Weizhi.Tang,k.nuamah,vbelle\}@ed.ac.uk} \\
}

\begin{document}

\ifcolmsubmission
\linenumbers
\fi

\maketitle

\begin{abstract}
While LLMs have demonstrated impressive abilities across various domains, they struggle with two major issues. The first is that LLMs trap themselves into local optima and the second is that they lack exhaustive coverage of the solution space. To investigate and improve these two issues, we propose \LR, a neuro-symbolic reasoning framework building on the integration of LLMs, genetic algorithms, and symbolic systems, comprising 7 essential components. Through conducting extensive experiments with 4 LLMs across 3 types of problems, we demonstrated the efficacy of \LR. Furthermore, with 7 additional ablation experiments, we further systematically analyzed and elucidated the factors that affect its performance. In addition, based on \LR, we extend the ideas to the fine-tuning process of LLMs and introduce \GF which enables a weaker model to imitate the reasoning process of a stronger model that reason under the \LR reasoning framework. We demonstrate that the significant effectiveness of \GF by conducting extensive experiments against 9 constructed baselines. We finally reveal the limitations and provide insights into future directions.
\end{abstract}

\section{Introduction}
\label{sec:lyria_introduction}

Large Language Models (LLMs) have demonstrated versatile abilities across various domains and tasks, benefiting from the large-scale corpora they are trained on~\citep{jiang2024surveylargelanguagemodels, llm-planning, pan-etal-2023-logic, tom-tang}. Nevertheless, their performance remains inferior, especially when faced with complex problems, which are typically characterized by their immense solution spaces, precise constraint satisfaction, multi-objective optimization, domain-specific prior knowledge, and so on, such as reasoning~\citep{mittal2025fcorebenchlargelanguagemodels}, planning~\citep{llm-planning}, theorem proving~\citep{song2025leancopilotlargelanguage}, code generation~\citep{jiang2024surveylargelanguagemodels}, etc. In this work, we primary consider two major issues that the LLMs encounter in their reasoning process given complex reasoning problems. The first is that LLMs trap themselves into local optima and the second is that they lack exhaustive coverage of the solution space. 

Genetic algorithms, a subset of evolutionary algorithms inspired by natural selection, powered by their essential operators such as selection, crossover, and mutation, are commonly used to approach optimal solution by iteratively optimizing the population through generations~\citep{Katoch_Chauhan_Kumar_2021,Gendreau_Potvin_2019,Koza_1994}. They have been studied and applied across diverse fields, such as reasoning~\citep{hameed2023optimized,schafer2015breeding,tamaddoni2001using}, planning~\citep{burns2024plancriticformalplanninghuman,path_planning}, combinatorial optimization~\citep{neural-ga-co,Kobler2009}, and symbolic regression~\citep{dual-objective-opt-neural-ga,Ashok2020LogicGG}, primarily due to their ability to escape local optima and conduct systematical searches, thereby enabling them to approach global optimal solutions~\citep{Katoch_Chauhan_Kumar_2021}. Therefore, integrating the genetic algorithm with LLMs is a promising and natural step to eliminate the first issue. 

However, for the second issue, one may argue that we could leverage Retrieval-Augmented Generation (RAG) to alleviate the issue by fetching additional information from external sources, thereby partially enriching the solution space~\citep{rag,gao2024retrievalaugmentedgenerationlargelanguage}. Nevertheless, when facing problems with an exponentially growing solution space, RAG normally fail to retrieve meaningful information since even external sources may not contain the solution space of the problems. For example, in the Traveling Salesman Problem (TSP), given $n$ cities, the size of its solution space reaches $\frac{(n-1)!}{2}$. This issue not only happens in problems like TSP, but also could happen in problems like code generation, planning, and so on. To eliminate this issue, we believe that leveraging symbolic systems, which typically have complete coverage of solution spaces, is a promising approach. 

Therefore, by leveraging both genetic algorithms and symbolic systems, we introduce a neuro-symbolic reasoning framework called \LR, consisting of 7 essential components, aiming to enhance the capability of LLMs to tackle complex reasoning problems. To evaluate its effectiveness, we select 3 types of combinatorial reasoning problems, for the reason that they are typically characterized by immense solution space and they are easily synthesized to avoid data pollution, and we conduct experiments using 4 LLMs on them against 2 constructed baselines, demonstrating its significant performance improvements. Furthermore, we also executed 7 additional ablation experiments to analyze and reveal the impact of various factors on its performance. In addition, based on \LR, we further propose \GF, a novel method which fine-tunes a weaker model to imitate the reasoning process of a stronger model that leverages \LR as its reasoning framework. By conducting extensive experiments, we show that \GF achieves significant performance improvements across 9 constructed baselines.

We summarize our contributions as follows:
\begin{enumerate}
\item We propose \LR, a neuro-symbolic reasoning framework by integrating LLMs with genetic algorithms and symbolic systems, for complex reasoning problems, and demonstrated its effectiveness through evaluations with 4 LLMs on 3 types of combinatorial reasoning problems;
\item We conduct 7 additional ablation experiments to comprehensively analyze the impact of various factors on the performance of \LR and demonstrate the indispensability of each component for \LR;
\item Based on \LR, we further propose \GF, a novel method to improve the performance of a weaker models by enabling it to imitate the reasoning process of a stronger model that reasons in the framework of \LR, and demonstrate that \GF can lead to consistent and significant performance improvements by conducting experiments against 9 constructed baselines.
\end{enumerate}

\section{Related Work}
\label{sec:lyria_related_work}
Recently, research on the integration of LLMs and genetic algorithms has begun to emerge, demonstrating promising results across a variety of tasks.

The integration of LLMs and genetic algorithms is used to tackle specific problems. Through synergistically combining LLMs with evolutionary algorithms, \textsc{EvoPrompt}~\citep{guo2024connectinglargelanguagemodels} shows its efficacy to optimize discrete prompt generation, outperforming existing automatic prompt generation methods across various LLMs and tasks. In addition, \citet{guilded_evo} propose a novel framework which leverages LLMs to autonomously evolve neural network architectures through feedback-driven code modifications via Evolution of Thought and Character Role Play, while \citet{llmatic} propose a method, LLMatic, that integrates the code-generation capabilities of LLMs and Quality Diversity algorithms which are a subset of evolutionary algorithms~\citep{cully2017quality,pugh2016quality} to efficiently discover network architectures. Furthermore, \citet{LLaMEA} propose LLaMEA, \citet{EoH} propose EoH, and \citet{ReEvo} propose ReEvo, to leverage LLMs to automatically construct and generate code of novel metaheuristic algorithms. Moreover, \citet{code-llm-gi} propose an approach based on LLMs and Genetic Improvement to improve code generation.  Furthermore, \citet{hemberg2024evolving} propose and demonstrate the way to replace traditional genetic programming operators with LLM-based operators to evolve code. In addition, \citet{tang-etal-2025-hygenar} introduce a hybrid LLM-driven genetic algorithm designed specifically to improve the ability of LLMs in the task of context-free grammar generation and demonstrated its significant effectiveness and potential. Nevertheless, these prior works differ from our proposed method, \LR, in either methodology or perspectives of analysis. Most existing works mainly leverage the integration of LLMs and genetic algorithms to tackle a specific problem, rather than proposing an LLM-based neuro-symbolic reasoning framework that integrates LLMs, genetic algorithms, and symbolic systems, to explore and eliminate the two main issues.

In addition, the integration of LLMs and genetic algorithms is also used in the process of fine-tuning LLMs. \citet{majumdar-etal-2025-genetic} introduce Genetic Instruct to improve the code generation ability of LLMs by using genetic algorithm to generate new code instructions given a small seed of initial population and fine-tuning models based on the new code instructions. Furthermore, \citet{qiu2025evolutionstrategiesscalellm} leverages Evolution Strategies (ES) to update parameters in the fine-tuning process and demonstrate that ES can be scaled up to fine-tune LLMs in billions parameter size and outperform existing Reinforcement Learning-based fine-tuning method in multiple aspects. In addition, \citet{han-etal-2025-attributes} propose a method called Genetic Prompt, which treats text semantic attributes as genes and using LLMs as genetic operators to synthesize data closer to real-world to fine-tune models, and demonstrate it can boost the model performance. However, although several prior works incorporate genetic algorithms in the fine-tuning process of LLMs, our proposed fine-tuning approach, \GF, remains fundamentally different. \GF is closely related to knowledge distillation, specifically reasoning patterns distillation, where a weaker model is trained to approach the reasoning behavior of a stronger model~\citep{xu2024surveyknowledgedistillationlarge}. Prior works proposes several reasoning patterns distillation methods~\citep{hsieh-etal-2023-distilling,zhu-etal-2024-pad,xu2024surveyknowledgedistillationlarge,chen2025-distilling-adaptive-thinking}. For example, \citet{hsieh-etal-2023-distilling} introduce a method that uses stronger LLMs to generate step-by-step rationales and then train smaller models with those rationales and labels. \citet{zhu-etal-2024-pad} propose Program-aided Distillation, which replaces CoT with programs, enabling automatic verification and filtering of incorrect reasoning processes, and then fine-tune smaller models on these reasoning processes data to improve reasoning performance of smaller models. However, distinct from previous reasoning patterns distillation methods, in \GF, a model is fine-tuned in the way of imitating the reasoning processes of a stronger model which operates under the proposed reasoning framework \LR, by which it not only enables smaller models imitate the reasoning processes of stronger models, but also mitigates the two issues mentioned before.

\section{Benchmarks}
\label{sec:lyria_benchmarks}
To evaluate \LR, we selected 3 types of combinatorial reasoning problems, i.e., Sudoku~\citep{sudoku_np}, Traveling Salesman Problem~\citep{tsp_np}, and Graph Coloring~\citep{gc_complexity}\footnote{We use the term \textit{combinatorial reasoning problems}, rather than using NP-hard, NP-complete, or combinatorial optimization problems, to emphasize their natural as complex reasoning problems in this work.}. All the 3 types of problems are characterized by their vast solution spaces and stringent constraints satisfaction requirements, thereby supposed to pose significant challenges to LLMs. It is worth noting that \LR is not restricted to these problems, and the selection of them is only motivated by their complexity and the convenience of generating uncontaminated data. We describe each problem, their metrics, and the challenges generation in the following sections.

\subsection{Sudoku}
Sudoku (SK) is a number-placement puzzle played on a $9\times 9$ grid, where each cell must be assigned a digit from 1 to 9. The grid is partitioned into nine $3\times 3$ subgrids. A correct Sudoku solution requires that each row, column, and subgrid contains all digits from 1 to 9 exactly once. The puzzle is typically presented with some cells pre-filled, and a given solution must respect the constraints. 

Let $s^\prime$ denote a SK solution, defined as a $9\times9$ matrix with each cell filled: 
$$
s^\prime = \begin{bmatrix}
u_{1,1} & u_{1,2} & u_{1,3} & \cdots & u_{1,9} \\
u_{2,1} & u_{2,2} & u_{2,3} & \cdots & u_{2,9} \\
u_{3,1} & u_{3,2} & u_{3,3} & \cdots & u_{3,9} \\
\vdots  & \vdots  & \vdots  & \ddots & \vdots  \\
u_{9,1} & u_{9,2} & u_{9,3} & \cdots & u_{9,9}
\end{bmatrix}.
$$
Let $\mathbb{S}$ denote the solutions to all SK problems. We mainly employ 3 metrics to assess the quality of the generated solutions: 

\begin{itemize}
    \item \textbf{Correctness:} The correctness of $s^\prime$ is defined as $\mathrm{CR(s^\prime)}$, which returns $1$ if $s^\prime$ satisfies all the row, column, and subgrids constraints, otherwise 0. Formally, let $R(s’,i)$ denote a set of values at row $i$, $C(s’,j)$ denote a set of values at column $j$, and $B(s’,k)$ denote a set of values in the $k$-th $3\times3$ subgrid. The $\mathrm{CR}(s^\prime)$ is given as:
\begin{align*}
\mathrm{CR}(s^\prime) =
\begin{cases}
1, & \text{if } (\forall i,\, |R(s^\prime,i)| = 9) 
    \land (\forall j,\, |C(s^\prime,j)| = 9)
    \land (\forall k,\, |B(s^\prime,k)| = 9)\\
    & \quad \; \land \; (\forall i,j,\, s^\prime[i,j] \in \{1,\dots,9\}) \\
0, & \text{otherwise}
\end{cases}
\end{align*}
We report the correctness percentage of $\mathbb{S}$ as $\mathrm{Sudoku}_{CR}$.
    \item \textbf{Score:} Evaluating a SK solution solely based on its correctness provides an overly narrow perspective, since when an LLM fails to provide a fully correct solution, it becomes challenging to observe and analyze whether it yields a high-quality albeit suboptimal solution and how closely it approximates the optimal solution. Thus, we define the score of a given $s^\prime$ as the average of percentages of constraints sanctification of rows, columns, and subgrids. Higher score means $s^\prime$ approaches more to the optimal solution. Formally, let $\mathbb{I}(X)$ be an indicator function that returns $1$ if $X = \{1,\dots,9\}$, otherwise 0. Let $\mathbb{I}^R_{s^\prime}=\sum_{i=1}^9\mathbb{I}(R(s^\prime,i))$, $\mathbb{I}^C_{s^\prime}=\sum_{j=1}^9\mathbb{I}(C(s^\prime,j))$, and $\mathbb{I}^B_{s^\prime}=\sum_{k=1}^9\mathbb{I}(B(s^\prime,k))$. We have $\mathrm{SC}(s^\prime)$ as:
$$
\mathrm{SC}(s^\prime) = 100 \cdot \frac{1}{27}\Bigl(\mathbb{I}^R_{s^\prime} + \mathbb{I}^C_{s^\prime} + \mathbb{I}^B_{s^\prime} \Bigr),
$$
We report the average of $\mathbb{S}$ as $\mathrm{Sudoku}_{SC}$.
    \item \textbf{Penalized Score:} To prevent a solution from resorting to extreme strategies to improve correctness, such as boosting row and column correctness while sacrificing subgrid correctness, and thus ending up far from the optimal solution despite a seemingly high score, we adopt the geometric mean to mitigate the impact of these extreme values on the overall score. Formally, we have $\mathrm{PS}(s^\prime)$ as:
$$
\mathrm{PS}(s^\prime) = 100 \cdot \sqrt[3]{\frac{\mathbb{I}^R_{s^\prime}}{9} \cdot \frac{\mathbb{I}^C_{s^\prime}}{9} \cdot \frac{\mathbb{I}^B_{s^\prime}}{9}},
$$
We report the average penalized score of $\mathbb{S}$ as $\mathrm{Sudoku}_{PS}$.
\end{itemize}

To generate a set of SK challenges for evaluation, we fixed the number of unfilled cells to 40 in each puzzle, generating a total of 50 distinct $9\times 9$ SK instances.

\subsection{Graph Coloring}
Graph Coloring (GC) is the task of assigning colors to the vertices of a given graph such that no two adjacent vertices share the same color. Formally, given a graph $G = (V,E)$ and a set of $k$ distinct colors, the goal is to find a function $f:V\rightarrow \{0,1,\dots,k-1\}$ such that for any edge $(u,v) \in E$, $f(u) \neq f(v)$.

Let $f^\prime$ be a GC solution. Let $\mathbb{F}$ denote the solutions to all GC problems. We mainly employ 4 metrics to evaluate the solution quality:
\begin{itemize}
    \item \textbf{Correctness:} The correctness of $f^\prime$ indicates whether it assigns colors to each vertex such that no adjacent vertex has the same color. Formally:
        \[
        \mathrm{CR}(f') =
        \begin{cases}
        1, & \text{if }
              \bigl(
                  \forall v\in V, f'(v)\in \{i\}_{i=0}^{k-1} 
              \bigr) \;\land
              \bigl(
                  \forall (u,v)\in E, f'(u)\neq f'(v)
              \bigr),\\
        0, & \text{otherwise.}
        \end{cases}
        \]
        We report the correctness percentage for $\mathbb{F}$ as $\mathrm{GC}_{CR}$.
    \item \textbf{Conflict Ratio}
        We define the conflict ratio as the ratio of edges whose endpoints share the same color. Define an indicator function:
        \begin{align*}
        \mathbb{I}_{CF}(f^\prime,u,v) &= 
        \begin{cases}
            1, & \text{if } f^\prime(u) = f^\prime(v),\\
            0, & \text{otherwise.}
        \end{cases}
        \end{align*}
        The conflict ratio $\mathrm{CF}$ for a given $f^\prime$ is given as:
        $$
        \mathrm{CF}(f^\prime) = \frac{ \sum_{(u,v)\,\in\,E} \mathbb{I}_{CF}(f^\prime,u,v)}{|E|}.
        $$
        We report the average of $\mathbb{F}$ as $\mathrm{GC}_{CF}$.
    \item \textbf{Score:} We define the score of a given $f^\prime$ as the percentage of edges colored properly as:
        $$
        \mathrm{Score}(f^\prime) = \left(1 - \mathrm{CF}(f^\prime) \right) \cdot 100.
        $$
        We report the average of $\mathbb{F}$ as $\mathrm{GC}_{SC}$.
    \item \textbf{Penalized Score:}
        The penalized score is a modified score that penalizes solutions using too many colors, e.g., coloring each vertices with a distinct color. Let \(k^\prime = \lvert \{f^\prime(v) : v \in V\}\rvert\) be the total number of distinct colors used. Formally, we have $\mathrm{PS}$ as:
        $$
        \mathrm{PS}(f^\prime)
        = 
        \begin{cases}
        0, & \text{if } k' \ge |V|, \\
        \mathrm{Score}(f^\prime), & \text{if } k' \le k, \\
        \mathrm{Score}(f^\prime)\,\cdot\, R, & \text{otherwise}.
        \end{cases}
        $$
        where $R = (1 - \dfrac{k' - k}{|V| - k})$ is the penalty ratio for the score. We report the average of $\mathbb{F}$ as $\mathrm{GC}_{PS}$.
\end{itemize}

To generate a set of GC challenges for evaluation, we fixed the number of vertices $|V|$ to $9$, the size of the color set $k$ to $3$ and the edge connection probability to $0.5$. This process yields $50$ distinct GC instances.

\subsection{Traveling Salesman Problem}
Traveling Salesman Problem (TSP) is a route-finding task defined on a set of cities and pairwise distances between them. Formally, let $G = (V,E)$ be a complete undirected graph in which $V=\{v_1,\dots,v_n\}$ is a set of vertices of the graph and $E = \{(u,v):u,v \in V, u\neq v\}$ is a set of edges. A distance function $d: V\times V \rightarrow \mathbb{R}_{\geq 0}$ assigns each edge $(u,v)$ a nonnegative distance $d(u,v)$. The goal in the TSP is to find a Hamiltonian cycle in $G$ whose total distance is minimized. Formally, a route $r$ is any permutation $\pi$ of $V$, with the first element also appearing at the end, forming a cycle, defined as a sequence $r = [v_{\pi(1)},\dots,v_{\pi(n)},v_{\pi(1)}]$. The total distance of this route is given as:
$
\mathrm{D}(r) = d(v_{\pi(n)},v_{\pi(1)}) + \sum^{n-1}_{i=1} d(v_{\pi(i)},v_{\pi(i+1)}).
$
Thus, the goal of TSP is to determine the route $r^*$ in all possible routes $R$ such that $\forall r \in R, \, \mathrm{D}(r) \geq \mathrm{D}(r^*)$, i.e., $r^* = \underset{\mathrm{r} \in R}{\min}\,\mathrm{D}(r)$.

Let  $r^\prime \in R$ be a TSP solution and $r^\prime = [v^\prime_1,v^\prime_2,\dots,v^\prime_{|V|+1}]$. Let $\mathbb{T}$ denote the solutions to all problems. We mainly employ 4 metrics to analyze the solution quality:
\begin{itemize}
    \item \textbf{Correctness:} The correctness of $r^\prime$ indicates whether it is the shortest route while starting and ending at $v^\prime_1$. Formally:
        \begin{align*}
        \mathrm{CR}(r^\prime) &= 
        \begin{cases}
        1, & \text{if } (r^\prime = \underset{r \in R}{\min}\,\mathrm{D(r)}) \\
           & \, \wedge \; (v^\prime_{1} = v^\prime_{|v|+1})\\
           & \, \wedge \; (|r^\prime| = |V| + 1)\\
        0, & \text{otherwise.}
        \end{cases}
        \end{align*}
        We report the average of $\mathbb{T}$ as $\mathrm{TSP}_{CR}$.
    \item \textbf{Excess Distance Multiplier:} The excess distance multiplier of $r^\prime$ quantifies the factor by which a solution’s distance exceeds the optimal distance $D(r^*)$. Formally:
        $$
        \mathrm{EDM}(r^\prime) = \min\Bigl(\,3,\frac{D(r^\prime) \;-\; D(r^*)}{D(r^*)}\Bigr)
        $$
        The value of 0 indicates the solution’s distance matches the optimal distance, while a value of 1 means the solution is twice as long as the option distance and so on. We make the value saturates at 3 as a default and maximum. We report the average of $\mathbb{T}$ as $\mathrm{TSP}_{EDM}$.
    \item \textbf{Missing Cities:} A given route $r^\prime$ may skip some cities or revisit others. To measure this, we count the number of distinct cities that are missed from the solution. Formally:
        $$
        \mathrm{MC}(r^\prime) = |V| - |\{v \mid v\in r^\prime \}|.
        $$
        We report the average of $\mathbb{T}$ as \(\mathrm{TSP}_{MC}\).
    \item \textbf{Penalized Score}
        The penalized score for $r^\prime$ measures how closely it approximates $r^*$, while applying penalties to the exceeded distance and also the omission of cities. Let $\mathrm{Dist} = 1-\frac{\mathrm{EDM}(r^\prime)}{3}$ and $\mathrm{Miss} = 1- \frac{\mathrm{MC(r^\prime)}}{|V|}$. Formally:
        $$
        \mathrm{PS}(r^\prime) = IS \cdot \min(\mathrm{Dist},\mathrm{Miss})
        $$
        where $IS = 100$ is the initialized score. We report the average of $\mathbb{T}$ as \(\mathrm{TSP}_{PS}\).
\end{itemize}

To generate a set of TSP challenges for evaluation, for each TSP problem, we fixed the number of cities $|V|$ to 10. The coordinate $(x_v,y_v)$ of city $v$ is sampled as $x,y \overset{i.i.d}{\sim} \mathcal{U}[0,100]$ and the distance between cities are calculated by Euclidean distance. A start and end city is fixed and noted as $v_{1}$. An optimal reference route is then derived by exhaustively enumerating all Hamiltonian cycles beginning and ending at $v_1$. This procedure is repeated to produce 50 distinct TSP instances.

\section{The \LR Reasoning Framework}
\label{sec:lyria_lyria}

\begin{figure}[t!]
\begin{center}
  \includegraphics[width=\columnwidth]{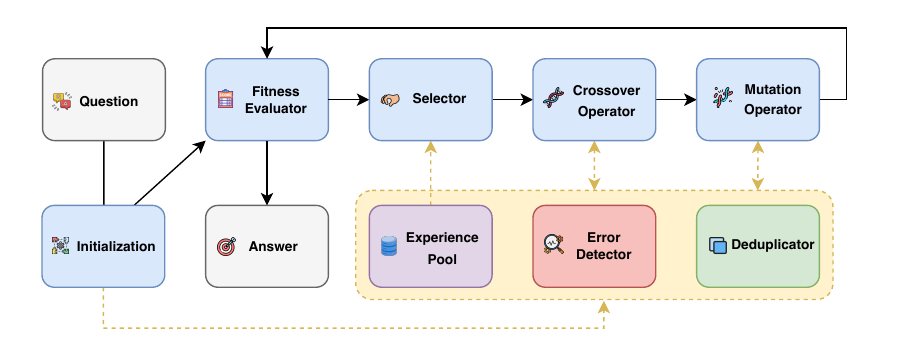}
  \caption{The \LR reasoning framework, consisting of 7 essential components, i.e., \textit{Error Detector}, \textit{Deduplicator}, \textit{Experience Pool},  \textit{Fitness Evaluator}, \textit{Selector}, \textit{Crossover Operator}, and \textit{Mutation Operator}, enables evolving candidate solutions through generations to obtain superior solution.}
  \label{fig:lyria_workflow}
\end{center}
\end{figure}

\subsection{Methodology}
\label{sec:lyria_methodology}
The \LR reasoning framework comprises 7 primary components: \emph{Error Detector}, \emph{Deduplicator}, \emph{Experience Pool}, \emph{Fitness Evaluator}, \emph{Selector}, \emph{Crossover Operator}, and \emph{Mutation Operator}. We begin with a high-level overview of the framework, followed by a detailed elucidation of each component in Section~\ref{sec:lyria_ed}, Section~\ref{sec:lyria_dd}, Section~\ref{sec:lyria_ep}, Section~\ref{sec:lyria_fe}, Section~\ref{sec:lyria_sl}, Section~\ref{sec:lyria_co}, Section~\ref{sec:lyria_mo}, respectively. Subsequently, we provide and explain the pseudo code in Section~\ref{sec:lyria_pseudo_code}.

Initially, an LLM generates a population of candidate solutions. Every candidate is scored by the \textit{Fitness Evaluator} and analyzed by the \textit{Error Detector}. Evolution then proceeds in generations: a fraction of the lowest fitness individuals, determined by the \textit{replay rate}, is replaced by the highest fitness candidates drawn from the \textit{Experience Pool}; the \textit{Selector} chooses appropriate parents; the \textit{Crossover Operator}, guided by parental errors, generates offspring until the population size is restored; and the \textit{Mutation Operator} modifies each candidate according to its own errors. After initialization and every crossover and mutation operations, the \textit{Deduplicator} removes duplicates to maintain diversity. The updated population is re‑evaluated and advanced in the next generation, until reaching the predetermined maximum number of generations. 

\subsubsection{Error Detector}
\label{sec:lyria_ed}
Inspired by Reflexion~\citep{shinn2023reflexion} and Self-Refinement~\citep{madaan2023selfrefine}, whenever a new candidate is generated, the error detector (ED) identifies its errors up to a predefined \textit{maximum detected errors}, enabling crossover and mutation operators to learn from past mistakes, thereby promoting the generation of improved candidates.  In \LR, we proposed two types of EDs: \textit{Verifier‑based} ED (VED) and \textit{LLM‑based} ED (LED). For both of the two types of EDs, they are mainly consisted of two error types, i.e., \textit{Syntax Error} (XE) which indicates whether the syntax or format of a given solution is correct, and \textit{Semantic Error} (SE) which reveals the information of the incorrect parts of the solution.

\paragraph{VED}
\label{sec:lyria_ved}
A VED invokes external symbolic systems, e.g., parsers, compilers, test suites, model checkers, etc., to examine a candidate against formal criteria and to emit deterministic and unbiased diagnoses. We designed specific VEDs for each problem type: 

\begin{itemize}
    \item \textbf{SK VED}: Let $s^\prime$ be a SK solution. With respect to the XE of SK, noted as $\mathrm{XE}_\mathrm{SK}$, for any generated SK solution by LLMs, we require it to be in the format of a $9\times 9$ matrix, with each cell separated by a space, as described in Prompt Template~\ref{pt:lyria_sudoku_dp}. If the format is correct, $\mathrm{XE}_\mathrm{SK}(s^\prime) = 0$, otherwise $\mathrm{XE}_\mathrm{SK}(s^\prime)=1$. With respect to the SE of SK, noted as $\mathrm{SE}_\mathrm{SK}$, it consists of indices of cells that do not satisfy its row, column, and subgrid constraints. We define it as: $\mathrm{SE}_\mathrm{SK}(s^\prime) = \{(i,j,t)\mid i,j \in \{1,\dots,9\} \,\land\, t \in \{ 0,1,2 \}\} \},$ where $i,j$ are row and column index respectively and $t$ indicates the unsatisfied constraint, in which $0$ indicates row, $1$ means column, and $2$ refers to subgrid.
    \item \textbf{GC VED}: Let $f^\prime$ be a GC solution, represented as a sequence $\mathcal{F^\prime} = [o_1^\prime,o^\prime_2,\dots,o^\prime_{|V|}]$ where $o_i = f^\prime(v_i)$ is a color assignment for $v_i \in V$. With respect to the XE of GC, noted as $\mathrm{XE}_\mathrm{GC}$, for any generated GC solution by LLMs, we require it to be in the format of a list of digits separated by commas, as described in Prompt Template~\ref{pt:lyria_gc_dp}. If the format is valid, $\mathrm{XE}_\mathrm{GC}(\mathcal{F}^\prime) = 0$, otherwise $1$. With respect to the SE of GC, noted as $\mathrm{SE}_\mathrm{GC}$, it is composed of two components: \textit{Conflict Edges} (CE) which consists of a set of edges where adjacent nodes share the same color, violating coloring constraints, i.e., $\mathrm{CE}(f^\prime) = \{(u,v) \mid (u,v) \in E \wedge f^\prime(u) = f^\prime(v)\}$, and \textit{Excess Colors Count} (ECC) which indicates the number of distinct colors used that exceeds the specified color count $k$, i.e., $\mathrm{ECC}(f^\prime) = |\mathcal{F}^\prime| - k$. Hence, $\mathrm{SE}_\mathrm{GC}$ for $f^\prime$ is given as: $\mathrm{SE}_\mathrm{GC}(f^\prime) = ( \mathrm{CE}(f^\prime), \mathrm{ECC}(f^\prime))$.
    \item \textbf{TSP VED}: Let $r^\prime$ be a TSP solution. With respect to the XE of TSP, noted as $\mathrm{XE}_\mathrm{TSP}$, for any generated TSP solution by LLMs, we require it to be in the format of a list of digits separated by commas, with the first and last city being the same and equal to 0 and each city index in the range from 0 to $number\_of\_cities - 1$, as described in Prompt Template~\ref{pt:lyria_tsp_dp}. If it is valid, $\mathrm{XE}_\mathrm{TSP}(r^\prime) = 0$, otherwise 1. With respect to the SE of TSP, noted as $\mathrm{SE}_\mathrm{TSP}$, it comprises two components: \textit{Missing Cities Set} (MCS) which consists of the cities omitted in $r^\prime$, i.e., $\mathrm{MCS}(r^\prime) = \{v \mid v \in V \land v \notin r^\prime\}$ and \textit{Excess Distance} (ED) which indicates the difference between the distance used in $r^\prime$ and $r^*$, i.e., $\mathrm{ED}(r^\prime) =D(r^\prime) - D(r^*)$. Hence, $\mathrm{SE}_\mathrm{TSP}$ is given as: $\mathrm{SE}_\mathrm{TSP}(r^\prime) = (\mathrm{MCS}(r^\prime),\mathrm{ED}(r^\prime))$.
\end{itemize}

\paragraph{LED}
An LED prompts an LLM to introspectively evaluate the candidate, harnessing its knowledge and reasoning abilities trained on corpora. By contrast to VED which always gives deterministic and robust errors information, LED may hallucinate on identifying errors and give imprecise error information. However, given a certain type of problem, in the case that external verifiers for this type of problem are unavailable and LLMs are capable of accurately capturing errors and providing errors information, LED remains indispensable. We crafted dedicated prompt templates for each of problem types by converting the rules of VEDs given in Section~\ref{sec:lyria_ved} to natural language, and demonstrated them in Prompt Template~\ref{pt:lyria_sudoku_ed},~\ref{pt:lyria_gc_ed}, and~\ref{pt:lyria_tsp_ed}. 

\subsubsection{Deduplicator}
\label{sec:lyria_dd}
During initialization, crossover, and mutation, the deduplicator (DD) discards any individual that duplicates an existing one, requesting replacements until a preset \textit{maximum deduplication attempts} is reached. This prevents identical candidates from dominating and preserves diversity. 

Formally, given a sequence of candidates $C = [c_i]_{i=1}^{k}$ where $k$ is the number of candidates generated, a newly generated candidate $c_{k+1}$, and a predefined \textit{maximum deduplication attempts} $\tau$, the deduplicator $\mathrm{DD}(C,c_{k+1}, \tau)$ operates as follows: if $c_{k+1} \notin C$, it returns $c_{k+1}$; if $\exists i \in \{1,\dots,\tau\},\, c^{(i)}_{k+1} \notin C$ and $\forall j < i,\, c^{(j)}_{k+1} \in C$, it returns $c^{(i)}_{k+1}$; otherwise, it returns $c^{(\mathcal{\tau})}_{k+1}$. Here, $c^{(i)}_{k+1}$ denotes a regenerated candidate. Hence, unique candidates are accepted immediately, and duplicates trigger up to $\tau$ regenerations, with the first non-duplicate retained or the final candidate accepted if all fail.

\subsubsection{Experience Pool}
\label{sec:lyria_ep}
During initialization and after each generation, candidate solutions with their fitness scores and errors are recorded in the experience pool (EP). Before selection, the lowest fitness individuals in the population are systematically replaced with the highest fitness candidates in EP, with the number of replacements determined by a predefined \textit{replay rate}. The EP preserves high-quality solutions and prevents inferior candidates from dominating the population, averting convergence toward suboptimal regions.

For the EP updating, formally, let $\mathrm{EP}_t$ denote EP at generation $t$, which is initialized as:
$
\mathrm{EP}_0 =
\left\{(c_i,s_i,e_i)\,\middle|\,
c_i \in C_0,\, s_i \in S_0,\, e_i \in E_0
\right\},
$
where $C_0=[c_i]_{i=1}^n$ is a sequence of candidates representing the initial population, $S_0=[s_i]_{i=1}^n$ is a sequence of fitness score corresponding to each candidate in $C_0$, and $E_0 = [e_1]_{i=1}^n$ is a sequence of error information of each candidate in $C_0$. After each generation $t$, the experience pool is updated as:
$
\mathrm{EP}_{t+1}= \mathrm{EP}_t \cup
\left\{(c_i,s_i,e_i)\,\middle|\,
c_i\in C_t,\, s_i\in S_t,\, e_i\in E_t
\right\}.
$

For the candidates replacement before selection, let $C_{t-1} = \{c_1,\dots,c_n\}$ be the previous population and $\mathrm{C}^{EP}_t = \{c^\star_1, \dots, c^\star_m\}$ be the candidates from EP at $t$. For the \textit{replay rate} $\rho$, we define replacement count $k = \lfloor \rho \cdot n \rfloor$. Let permutation $\sigma^\uparrow$ sort $C_{t-1}$ such that $s_{\sigma^\uparrow (i)} \leq s_{\sigma^\uparrow (j)}$ for all $i < j$, and permutation $\sigma^\downarrow$ sort $\mathrm{C}^{EP}_t$ such that $s^\star_{\sigma^\downarrow (i)} \geq s^\star_{\sigma^\downarrow (j)}$ for all $i < j$. We construct the new population $C' = [c^\prime_1,c^\prime_2,\dots,c^\prime_n]$, in which $c'_i = c^\star_{\sigma^\downarrow (i)}$ if $1 \leq i \leq k$ and $s^\star_{\sigma^\downarrow (i)} > s_{\sigma^\uparrow (i)}$, otherwise $c'_i=c_{\sigma^\uparrow (i)}$.

\subsubsection{Fitness Evaluator}
\label{sec:lyria_fe}
The fitness evaluator (FE) assigns each candidate a score, determining selection probability during evolution and influencing their crossover and mutation opportunities. Higher scores indicate the proximity of solutions to the optimum, increasing the possibility of selection to generate offspring. Lower scores reduce their selection probability. The FE critically guides the evolution and different FEs can exert distinct evolutionary processes. In \LR, we propose two types of FEs, i.e., Oracle-based FE and LLM-based FE.

\paragraph{Oracle-Based FE}
The Oracle‑based FE leverages an external symbolic system as a verifier that deterministically returns a score for a candidate solution. Given a candidate and the associated scoring criteria, the verifier strictly adheres to the criteria, producing a precise score. For all 3 problem types, we implement their \textit{penalized score} metric as the FE criteria in their verifiers to compute their fitness scores.

\paragraph{LLM-Based FE}
The LLM‑based FE eliminates the need for external or handcrafted verifiers by prompting an LLM to generate scores. Although an LLM can occasionally deviate from the ground‑truth score, it is still worthy and indispensable, when an external verifier is unavailable or costly to obtain. For each type of problem, we instruct the LLM to compute the fitness score based on their \emph{penalized score} metric. We demonstrate all prompt templates in Prompt Template~\ref{pt:lyria_sudoku_llm_based_fitness_evaluator},~\ref{pt:lyria_gc_llm_based_fitness_evaluator}, and~\ref{pt:lyria_tsp_llm_based_fitness_evaluator}.

\subsubsection{Selector}
\label{sec:lyria_sl}
Given a sequence of candidate solutions, the selector elects appropriate candidates and prepares a mating pool for subsequent crossover and mutation. While prioritizing individuals with high fitness may promote the generation of superior offspring, exclusively retaining them may impede population diversity and risk premature convergence to local optima. To balance exploration and exploitation, we implement a hybrid selection strategy for the selector that combines truncation selection with tournament selection. Nevertheless, we note that \LR is not confined to a specific selection strategy and it can be substituted by any alternatives.

Let $C = [c_i]_{i=1}^n$ be a sequence of candidates with fitness scores $S = [s_i]_{i=1}^n$. Let $k_e \in [0,n]$ denote the number of fittest candidates that are directly carried forward in truncation selection. Let $k_r = n-k_e$ denote the number of candidates for tournament selection. The selector sorts $C$ in descending order based on their fitness, giving 
$
C_{\sigma^\downarrow} = [c_{\sigma^\downarrow(1)},c_{\sigma^\downarrow(2)},\dots,c_{\sigma^\downarrow(n)}].
$ 
Then, it selects a subsequence of $k_e$ fittest candidates from $C_{\sigma^\downarrow}$, corresponding to the truncation selection, as 
$
C_{\mathrm{trunc}} = [c_{\sigma^\downarrow(1)},c_{\sigma^\downarrow(2)},\dots,c_{\sigma^\downarrow(k_e)}].
$
Then, let $I_{\mathrm{tour}} = [(x_i,y_i)]^{k_r}_{i=1}$ be a sequence of index-pairs and \(x_i, y_i \overset{\text{i.i.d.}}{\sim} \mathcal{U}[1, n]\). The candidates selected by tournament selection is given as 
$
C_{\mathrm{tour}} = [c^{\mathrm{tour}}_1,c^{\mathrm{tour}}_2,\dots,c^{\mathrm{tour}}_{k_r}],
$ 
where $c^{\mathrm{tour}}_i = c_{x_i}$ if $s_{x_i} > s_{y_i}$, otherwise $c^{\mathrm{tour}}_i = c_{y_i}$. Thus, combining the truncation selection and tournament selection, the selector is defined as:
$
\mathrm{Select}(C,S,k_e,k_r) = C_{\mathrm{trunc}} \cup C_{\mathrm{tour}}.
$
Hence, the candidates selected as the mating pool is $C^\prime = \mathrm{Select}(C,S,k_e,k_r)$.

\subsubsection{Crossover Operator}
\label{sec:lyria_co}
The crossover operator (CO) selects parent pairs from the mating pool, governed by a predefined \emph{crossover rate}. If crossover is skipped, a parent is randomly returned; otherwise, offspring are generated. This iterates until the offspring population reaches the predefined population size. The objective of CO is to combine advantageous traits of parents while suppressing detrimental ones to produce improved offspring with higher fitness. In \LR, we propose two COs, i.e. External CO (ECO) and LLM-based CO (LCO) , which are alternated in evolution based on an \emph{external crossover rate}, for which higher prioritizes ECO while lower prioritizes LCO. 

\paragraph{ECO} 
In ECO, two parent candidates are combined via external symbolic systems, based on domain-specific strategies and also the error information of the given parents. Varying from domains, distinct strategies can be employed, e.g., leveraging external heuristics provided by domain experts, formal logical constraints, etc. We designed specific ECOs for each problem type:

\begin{itemize}
    \item \textbf{SK ECO} Let $c_1, c_2$ be parent candidates as two $9\times9$ matrix. Let $P_{\mathrm{removed}}$ denote the initially removed positions of the puzzle and $\{ (i,j) \mid (i,j,t) \in \mathrm{SE}_\mathrm{SK}(c)\} \subseteq P_{\mathrm{removed}}$. The crossover operator $\mathrm{CO}_\mathrm{SK}(c_1,c_2)$ produces the child $c_\mathrm{child}$ as:  
\[
c_{\mathrm{child}} = 
\begin{cases} 
c_1 & \text{if } \mathrm{SE}_\mathrm{SK}(c_1) = \emptyset, \\ 
c_2 & \text{if } \mathrm{SE}_\mathrm{SK}(c_2) = \emptyset, \\ 
\Phi(c_2, P_{\mathrm{corr}}(c_1)) & \text{otherwise},
\end{cases}
\]  
where \( P_{\mathrm{corr}}(c_1) = P_{\mathrm{removed}} \setminus \{ (i,j) \mid (i,j,t) \in \mathrm{SE}_\mathrm{SK}(c_1)\} \) are \( c_1 \)’s corrected positions, and \( \Phi(c, P_{\mathrm{corr}}) \) updates \( c \) by replacing \( c \)’s values at \( \{ (i,j) \mid (i,j,t) \in \mathrm{SE}_\mathrm{SK}(c)\} \cap P_{\mathrm{corr}} \).
    \item \textbf{GC ECO} Let $c_1, c_2$ be two parent candidates, which are two GC solutions of which each represented as a sequence of color assignments, i.e., $c=[o_1,o_2,\dots,o_{|V|}]$ where $o_i$ is a color assignment for $v_i \in V$. Let $\mathrm{CV}(c) = \{v \mid (v,u) \in \mathrm{CE}(c)\}$ denote conflict vertices in candidate $c$. The external crossover operator $\mathrm{ECO}_\mathrm{GC}$ generates the child $c_{\mathrm{child}}$ as:  
\[
\forall i, c_{\mathrm{child}}[i] = 
\begin{cases} 
c_2[i] & \text{if } i \in \mathrm{CV}(c_1) \setminus \mathrm{CV}(c_2), \\ 
c_1[i] & \text{if } i \in \mathrm{CV}(c_2) \setminus \mathrm{CV}(c_1), \\ 
\mathrm{R}(c_1[i], c_2[i]) & \text{otherwise},
\end{cases}
\]  
where $i \in \{1,\dots, |V|\}$ indicates both an index in a color assignment sequence and also a city, and \( \mathrm{R}(x,y) \) denotes uniform random selection between \( x \) and \( y \). Thus, $\mathrm{CO}_\mathrm{GC}$ transfers non-conflicting colors between parents while preserving valid vertex assignments.
    \item \textbf{TSP ECO} Let \( c_1, c_2 \) be two parent candidates, which are two routines, for which each routine is a sequence of visited cities. The external crossover operator \( \mathrm{ECO}_\mathrm{TSP} \) generates the child $c_\mathrm{child}$ as:  
\[
c_{\mathrm{child}} = 
\begin{cases} 
c_1 & \text{if } \mathrm{SE}_\mathrm{TSP}(c_1) = (\emptyset, 0), \\ 
c_2 & \text{if } \mathrm{SE}_\mathrm{TSP}(c_2) = (\emptyset, 0), \\ 
\Psi(c_1, c_2, k) & \text{otherwise},
\end{cases}
\]  
where \( k \overset{\text{i.i.d.}}{\sim} \mathcal{U}[1, n-1] \) is a uniformly random crossover point, and  
\[
\forall i, \Psi(c_1, c_2, k)[i] = 
\begin{cases} 
c_1[i] & \text{for } i \leq k, \\ 
c_2[i] & \text{for } i > k,
\end{cases}
\]  
where $i \in \{1,\dots, |V|\}$ is an index of a sequence.
Thus, $\mathrm{CO}_\mathrm{TSP}$ merges partial routes from both parents while preserving order.
\end{itemize}

\paragraph{LCO} 
In LCO, an LLM is prompted to merge two parent candidates by integrating their advantageous attributes and excluding their deficiencies, drawing on the experience and prior knowledge of the LLM and their understanding of the error information of the candidates provided by the ED, to produce an improved child. This approach eliminates the need to manually specify any domain‐specific strategy, instead fully delegating it to the LLM. We demonstrated the prompts we designed for each of the 3 problem types in Prompt Template~\ref{pt:lyria_sudoku_lco},~\ref{pt:lyria_gc_lco}, and~\ref{pt:lyria_tsp_lco}.

\subsubsection{Mutation Operator}
\label{sec:lyria_mo}
The mutation operator (MO) applies mutations to each candidate based on a predefined \emph{mutation rate}, returning either the original or mutated candidate. This preserves population diversity and prevents premature convergence. In \LR, we propose two MOs, i.e., External MO (EMO) and LLM-based MO (LMO), which are alternated in evolution via an \emph{external mutation rate}, for which higher prioritizes EMO while lower prioritizes LMO.

\paragraph{EMO}
In EMO, the mutation is handled by external symbolic systems, guided by domain-specific mutation strategies and also the error information of the candidate, to modify the candidate. We designed specific EMOs for each problem type:

\begin{itemize}
    \item \textbf{SK EMO} Given a candidate $c$ as a $9\times 9$ matrix, the \( \mathrm{EMO}_\mathrm{SK} \) is given as:  
\[
\mathrm{EMO}_\mathrm{SK}(c) = 
\begin{cases} 
c & \text{if } \mathrm{SE}_\mathrm{SK}(c) = \emptyset, \\
\Theta(c, p, v) & \text{otherwise},
\end{cases}
\]  
where \( p \overset{i.i.d.}{\sim} \mathcal{U}(\{ (i,j) \mid (i,j,t) \in \mathrm{SE}(c)\}) \) is an error position, \( v \overset{i.i.d.}{\sim} \mathcal{U}[1, 9] \) is a random value, and \( \Theta(c, p, v) \) denotes replacing \( c \)’s value at the position \( p \) with the value \( v \).  
    \item \textbf{GC EMO} Given a candidate \( c \) as a sequence of color assignments, the \( \mathrm{EMO}_{\mathrm{GC}} \) proceeds as:  
\[
\mathrm{EMO}_{\mathrm{GC}}(c) = \Gamma(\Lambda(c)).
\]  
It is composed of two main components, i.e., Conflict Edges Resolution and Excess Colors Correction, denoted as $\Lambda$ and $\Gamma$ respectively. 

Conflict Edges Resolution is given as:
$$
\Lambda(c)[i] = 
\begin{cases} 
y  & \text{if } c[i] \in \mathrm{CV}(c)\\
c[i] & \text{otherwise},
\end{cases}
$$
where $i \overset{i.i.d.}{\sim} \mathcal{U}[1,|c|]$ is a randomly selected index, and $y \overset{i.i.d.}{\sim} \mathcal{U}(\{0,\dots,k-1\}\setminus \{c[i]\})$ is a new color value. Hence, it reassigns a new color to a vertex if it is conflicted such that $y \neq c[i]$. 
 
Excess Colors Correction is given as:
$$
\forall i, \Gamma(c)[i] = 
\begin{cases} 
z & \text{if } c[i] \in O, \\ 
c[i] & \text{otherwise}. 
\end{cases}
$$
where $i \in \{1,\dots,|c|\}$, $z \overset{i.i.d.}{\sim} \mathcal{U}(\{0,\dots,k-1\})$ and $O=\{k,\dots,k+\mathrm{ECC}(c)\}$. Hence, it replaces all color values $\geq k$ with random valid colors $z \in \{0,\dots,k-1\}$. 
    \item \textbf{TSP EMO} Give a candidate $c$ as a route, which is a sequence of visited cities, let $Y$ be a set of duplicated cities in $c$, and let $M = \mathrm{MCS}(c)$ be a set of missing cities, the \( \mathrm{EMO}_{\mathrm{TSP}} \) is given as:
\[
\mathrm{EMO}_{\mathrm{TSP}}(c) =
\begin{cases}
c & \text{if } M = \emptyset,\\
\Omega(c)  & \text{otherwise},
\end{cases}
\]
where \( \Omega \) is defined as:
\[
\forall i, \Omega(c)[i] =
\begin{cases}
\varphi(i) & \text{if } c[i] \in Y\\
c[i] & \text{otherwise},
\end{cases}
\]
in which $i \in \{1,\dots, |c|\}$, $\varphi$ is an injection from the first $r=\min(|Y|,|M|)$ elements of $Y$ into $M$. Hence, it resolves errors by substituting duplicates with missing cities, ensuring the mutated child becomes a route without missing cities.
\end{itemize}

\paragraph{LMO}
In LMO, an LLM is instructed to mutate a given candidate, by identifying and improving its inferior parts, based on the knowledge of LLMs and their understanding of error information. Similar to LCO, this approach eliminates the necessity to manually construct domain-specific strategies and enables the LLM to autonomously design strategies to modify the given candidate. We designed the prompt for each problem type and demonstrated them in Prompt Template~\ref{pt:lyria_sudoku_lmo},~\ref{pt:lyria_gc_lmo}, and~\ref{pt:lyria_tsp_lmo}.

\subsubsection{Pseudo Code}
\label{sec:lyria_pseudo_code}
The pseudo code of \LR is demonstrated in Algorithm~\ref{algo:lyria}, in which $\mathcal{P}$ means the problem, $n_p$ means \textit{population size}, $n_g$ means \textit{generations}, $k_e$ means the number of fittest candidates that are directly carried forward in truncation selection, $\epsilon$ means the \textit{maximum detected errors}, $\rho$ means \textit{replay rate}, $\tau$ means the \textit{maximum deduplication attempts}, $\eta$ means \textit{crossover rate}, $\xi$ means \textit{external crossover rate}, $\kappa$ means \textit{mutation rate}, $\mu$ means \textit{external mutation rate}, $\textsc{FE}$ means the fitness evaluator which can either be Oracle-based or LLM-based, $\textsc{ED}$ means the error detector which can either be Verifier-based or LLM-based, $\textsc{LCO}$ means the LLM-based crossover operator, \textsc{ECO} means the external crossover operator, \textsc{LMO} means the LLM-based mutation operator, \textsc{EMO} means the external mutation operator, and $llm$ indicates the LLM which accepts a prompt and returns a response\footnote{In Algorithm~\ref{algo:lyria}, "$||$" means concatenating two arrays.}.

\begin{algorithm}
\caption{Pseudo Code of \LR}
\label{algo:lyria}
{\singlespace
{\small
\begin{algorithmic}[1]
\Procedure{\LR}{$\mathcal{P}, n_p, n_g, k_e, \epsilon, \rho, \tau, \eta, \xi, \kappa, \mu, \textsc{FE}, \textsc{ED}, \textsc{LCO}, \textsc{ECO}, \textsc{LMO}, \textsc{EMO}, llm$}
    
    \State $population,fitness,best\_fitness,best\_solution, errors, \textsc{EP}, \tau^\prime \gets [], [], -\infty, \emptyset, [], \emptyset, 0$
    
    \While{$|population| < n_p $} \Comment{Initialization}
        \State $prompt \gets \textsc{DPPrompt}(\mathcal{P})$
        \State $candidate \gets llm(prompt)$
        \State \textbf{if} $child \in population$ \text{and} $\tau^\prime < \tau$ \textbf{then} $\tau^\prime \gets \tau^\prime + 1$; Continue \textbf{else} $\tau^\prime \gets 0$ \Comment{Deduplicator}
        \State $population,fitness \gets population \parallel [candidate], fitness \parallel [\textsc{FE}(candidate)]$
        \State $errors \gets errors \parallel [\textsc{ED}(candidate,\epsilon)]$
        \If{$fitness[-1] > best\_fitness$}
            \State $best\_solution, best\_fitness \gets candidate, fitness[-1]$
        \EndIf
    \EndWhile
    
    \For{$n_g^\prime \gets 1$ \textbf{to} $n_g$} \Comment{Start Evolution}
        \State $population, fitness, errors \gets$ Replacing $\lfloor n_p \cdot \rho \rfloor$ candidates with \textsc{EP} \Comment{EP Replay}
        
        \State $selected \gets \textsc{Select}(population, fitness, k_e, |population|-k_e)$ \Comment{Selection Phase}
        \State Update $fitness,errors$ to align with $selected$
        
        \State $offspring, os\_fitness, os\_errors, \tau^\prime  \gets [],[],[],0$
        \While{$|offspring| < n_p$} \Comment{Crossover Phase}
            \State $c_1, c_2, s_1, s_2, e_1, e_2 \gets \textsc{RandomChoice}(selected, fitness, errors)$
            \If{$\textsc{RD}() < \eta$} \Comment{Apply ECO or LCO , and $\textsc{RD}()$ means $x \overset{i.i.d.}{\sim}\mathcal{U}[0,1]$} 
                \State $child \gets$ \textsc{ECO}($\mathcal{P}, c_1, c_2, e_1, e_2$) \textbf{if} $\textsc{RD}() < \xi$ \textbf{else} \textsc{LCO}($\mathcal{P}, c_1, c_2, s_1, s_2, e_1, e_2, llm$)
            \Else
                \State $child \gets \textsc{RandomChoice}([c_1, c_2])$
            \EndIf
            \State \textbf{if} $child \in offspring$ \text{and} $\tau^\prime < \tau$ \textbf{then} $\tau^\prime \gets \tau^\prime + 1$; Continue \textbf{else} $\tau^\prime \gets 0$
            \State $offspring, os\_fitness \gets offspring \parallel [child], os\_fitness \parallel [FE(child)]$
            \State $os\_errors \gets os\_errors \parallel [ED(child,\epsilon)]$
        \EndWhile

        \State $mutated, mt\_fitness, mt\_errors, \tau^\prime \gets [],[],[], 0$
        \While{$|mutated| < n_p$}  \Comment{Mutation Phase}
            \State $i \gets |mutated|$
            \State $c, e, s \gets offspring[i], errors[i], fitness[i]$
            \If{$\textsc{RD}() < \kappa$} \Comment{Apply EMO or LMO}
                \State $child \gets$ \textsc{EMO}($\mathcal{P}, c, e$) \textbf{if} $\textsc{RD}() < \mu$ \textbf{else} \textsc{LMO}($\mathcal{P}, c, s, e, llm$)
            \Else
                \State $child \gets c$
            \EndIf
            \State \textbf{if} $child \in mutated$ \text{and} $\tau^\prime < \tau$ \textbf{then} $\tau^\prime \gets \tau^\prime + 1$; Continue \textbf{else} $\tau^\prime \gets 0$
            \State $mutated, mt\_fitness \gets mutated \parallel [child], mt\_fitness \parallel [FE(child)]$
            \State $mt\_errors \gets mt\_errors \parallel [ED(child,\epsilon)]$
        \EndWhile
        
        \State $population, fitness \gets mutated, mt\_fitness$ \Comment{Update Population}
        \State $errors \gets mt\_errors$
        \State $\textsc{EP} \gets \textsc{EP} \cup \{(c,s,e) \mid c \in population, s \in fitness, e \in errors \}$
        \State Update $best\_solution$ and $best\_fitness$
    \EndFor
    \State \Return $best\_solution$
\EndProcedure
\end{algorithmic}
}
}
\end{algorithm}

We provide a line-by-line explanation of the pseudo code as follows. At Line 2, we initialize essential parameters. From Line 3 to 12, we use $\textsc{DPPrompt}$ to construct the prompt. For example, we use prompt templates shown in Prompt Template~\ref{pt:lyria_sudoku_dp},~\ref{pt:lyria_gc_dp}, and~\ref{pt:lyria_tsp_dp} to construct the prompt for each problem type. Then, $llm$ uses this prompt to generate new candidate. Deduplicator removes redundant candidates, $\textsc{FE}$ assigns fitness score to the candidate, $\textsc{ED}$ detect its errors up to $\epsilon$ for which we show the prompt templates we used in Prompt Templates~\ref{pt:lyria_sudoku_ed},~\ref{pt:lyria_gc_ed}, and~\ref{pt:lyria_tsp_ed}. For Line 13-46, \LR starts evolution. At Line 14, EP intervenes to replace the lowest fitness candidates from the previous population with the highest fitness candidates from EP along with their fitness scores and errors. At Line 15-16, the selector selects appropriate candidates along with their fitness scores and errors. For Line 17-28, the crossover rate $\eta$ determines whether to apply the crossover operation, while the external crossover rate $\xi$ determines whether $\textsc{ECO}$ is prioritized. For $\textsc{LCO}$, we demonstrate the prompt templates we used for each problem type in Prompt Template~\ref{pt:lyria_sudoku_lco},~\ref{pt:lyria_gc_lco}, and~\ref{pt:lyria_tsp_lco}. For Line 29-41, the mutation rate $\kappa$ determines whether to apply the mutation operation, while the external mutation rate $\mu$ determines whether $\textsc{EMO}$ is prioritized. For $\textsc{LMO}$, we demonstrate the prompt templates we used for each problem type in Prompt Template~\ref{pt:lyria_sudoku_lmo},~\ref{pt:lyria_gc_lmo}, and~\ref{pt:lyria_tsp_lmo}. For Line 42-45, $population$, $fitness$, $errors$, and $\textsc{EP}$ are updated, and the evolution advances into the next generation.

In addition, in practice, to prevent unnecessary additional overhead, especially when $best\_fitenss$ attains its optima, we also introduce a fitness threshold. During both initialization and the end of each generation, we check whether $best\_fitness$ meets or exceeds the fitness threshold. The $best\_solution$ is returned earlier if it meets the threshold. For every type of problem, the fitness threshold is set to $100$.

\subsection{Main Experiment}
\label{sec:lyria_main_experiment}

\subsubsection{Baselines}
\label{sec:lyria_main_experiment_baselines}
We adapted two baselines, i.e., Direct Prompting (DP) and Best‑of‑N Direct Prompting (BoN), for comparative evaluation. We elaborate on the details and necessity of the two baselines as follows. A discussion of why comparisons with other LLM-driven genetic algorithms or metaheuristic algorithms are not included is provided in Section~\ref{sec:lyria_lim_fut}.

For DP, an LLM is invoked once and prompted to generate a solution directly in a zero-shot manner, in which it is required to think and reason in any way it deems appropriate before producing the final answer.  We designed the prompt templates for each problem type and demonstrated them in Prompt Template~\ref{pt:lyria_sudoku_dp},~\ref{pt:lyria_gc_dp}, and~\ref{pt:lyria_tsp_dp}.

Since \LR may sample an LLM up to $L$ times for a single problem, a naive comparison against DP could favor \LR merely by virtue of increased sampling\footnote{The calculation of $L$ is demonstrated in Appendix~\ref{appendix:lyria_l_calculation}.}. To ensure a fair comparison, we adopt the BoN approach. For each problem, BoN draws $N = L$ independent responses from the LLM using the identical prompt template employed by DP and preserves the one with the best metrics as the answer. Additionally, same as \LR, a deduplicator is introduced to remove redundant answers. This approach equalizes the number of sampling and ensures any observed performance improvements are attributable to the innovations of \LR.

\subsubsection{Experiment Settings}
\label{sec:lyria_main_experiment_settings}
For a comprehensive evaluation, we selected 4 LLMs: \textit{GPT-4o-Mini}, \textit{Qwen2.5:32B-Instruct}, \textit{Qwen2.5:7B-Instruct}, and \textit{Mistral:7B-Instruct}.

For DP, we set the \textit{temperature} to $0$ for greedy decoding and the \textit{maximum generated tokens} to $4096$.

For BoN, we set the \textit{temperature} at $0.7$ to enable diverse generated answers, the \textit{maximum generated tokens} at $4096$, the \textit{sampling times} $N$ at $345$ to align the number of queries with \LR, and the \textit{maximum deduplication attempts} at $3$.

For \LR, we switch on the Oracle-based FE. We set the \textit{temperature} of LLM at $0.7$, the \textit{maximum generated tokens} at $4096$, the \textit{population size} at $30$, the \textit{generations} at $15$, the \textit{maximum detected errors} at $3$, the \textit{maximum deduplication attempts} at $3$, the \textit{replay rate} at $0.6$, the \textit{crossover rate} at $0.7$, the \textit{external crossover rate} at $0.3$, the \textit{mutation rate} at $0.3$, and the \textit{external mutation rate} at $0.3$.

\subsubsection{Results \& Analysis}

\begin{table*}[t!]
\centering
\begin{tabularx}{\textwidth}{l c | X X | X X | X X}
\toprule
\textbf{Model} & \textbf{Method}
  & \textbf{$\mathrm{SK}_{CR}$} & \textbf{$\mathrm{SK}_{PS}$}
  & \textbf{$\mathrm{GC}_{CR}$}     & \textbf{$\mathrm{GC}_{PS}$}
  & \textbf{$\mathrm{TSP}_{CR}$}    & \textbf{$\mathrm{TSP}_{PS}$}\\
\midrule
\multirow{3}{*}{\centering GPT-4o-Mini}
& DP 
& 0 
& 39
& 0 
& 73
& 0 
& 79 \\
& BoN  
& 6
& 73
& 0
& 86
& 4
& 94 \\
& \LR
& \textbf{8}
& \textbf{73}
& 0
& \textbf{97}
& \textbf{6}
& \textbf{96} \\
\midrule
\multirow{3}{*}{\centering Qwen2.5:32B-Instruct}
& DP 
& 0 
& 31
& 0 
& 74
& 0 
& 81  \\
& BoN  
& 8
& 76
& 0
& 87
& 8
& 97 \\
& \LR
& \textbf{32}
& \textbf{87}
& 0
& \textbf{96}
& \textbf{30}
& \textbf{99} \\
\midrule
\multirow{3}{*}{\centering Mistral:7B-Instruct}
& DP 
& 0 
& 0
& 0 
& 0 
& 0 
& 60 \\
& BoN  
& 0
& 5
& 0
& 84
& 0
& 80 \\
& \LR
& 0
& \textbf{12}
& 0
& \textbf{92}
& 0
& \textbf{89} \\
\midrule
\multirow{3}{*}{\centering Qwen2.5:7B-Instruct}
& DP 
& 0 
& 26 
& 0 
& 73 
& 0 
& 34  \\
& BoN  
& 0
& 55
& 0
& 84
& 0
& 88\\
& \LR
& 0
& \textbf{61}
& 0
& \textbf{95}
& \textbf{4}
& \textbf{95} \\
\bottomrule
\end{tabularx}
\caption{The results of Correctness and Penalized Score for SK, GC, TSP. For each LLM, across the three methods, the best correctness and penalized score are highlighted with a bold font.}
\label{tab:lyria_main_results}
\end{table*}

As shown in Table~\ref{tab:lyria_main_results}, LLMs struggle across problems with DP\footnote{Results of all metrics for each problem type are shown in Appendix~\ref{appendix:lyria_additional_results}.}. While BoN greatly improves the performance across the problems, \LR demonstrates its ability to further consistently contribute significant improvement across various LLMs and problems. For example, \LR improves $\mathrm{GC}_{PS}$ for \textit{GPT-4o-Mini} by $24$ over DP and $11$ over BoN, while enhancing $\mathrm{SK}_{CR}$ by $32\%$ and $24\%$ and also $\mathrm{SK}_{PS}$ by $56$ and $11$, for \textit{Qwen2.5:32B-Instruct}, compared to DP and BoN, respectively. In addition, for relatively small LLMs like \textit{Qwen2.5:7B-Instruct}, \LR also shows its efficacy, by $22$ and $11$ $\mathrm{GC}_{PS}$ increases, and $61$ and $7$ $\mathrm{TSP}_{PS}$ improved, compared to DP and BoN, respectively. 

Furthermore, across all LLMs, for SK, \LR shows an average $10\%$ and $7\%$ increases on $\mathrm{SK}_{CR}$ with $34$ and $6$ increases on $\mathrm{SK}_{PS}$, compared to DP and BoN. For GC, \LR shows $40$ and $10$ improvement on $\mathrm{GC}_{PS}$. For TSP, \LR shows $10\%$ and $7\%$ increases on $\mathrm{TSP}_{CR}$ with $32$ and $5$ improvement on $\mathrm{TSP}_{PS}$. Therefore, across all LLMs and problems, \LR demonstrates $7\%$ and $5\%$ increases on the correctness and $35$ and $7$ improvements on the penalized score, compared to DP and BoN, respectively, demonstrating the consistent performance contribution offered by \LR.

\subsection{Ablation Experiments}
\label{sec:lyria_ablation_experiments}
To further investigate the impact of various factors that influence the performance of \LR, we conducted 7 additional experiments. To avoid prohibitive costs, we selected \textit{Qwen2.5:7B-Instruct} and limited the number of problems to $10$. Unless otherwise specified, we adhere to the same parameter settings as in the main experiment and refer their performance to the \textit{penalized score} metric.

\subsubsection{Scaling Population Size and Generations}

\begin{figure}[t!]
  \begin{center}
  \includegraphics[width=\columnwidth]{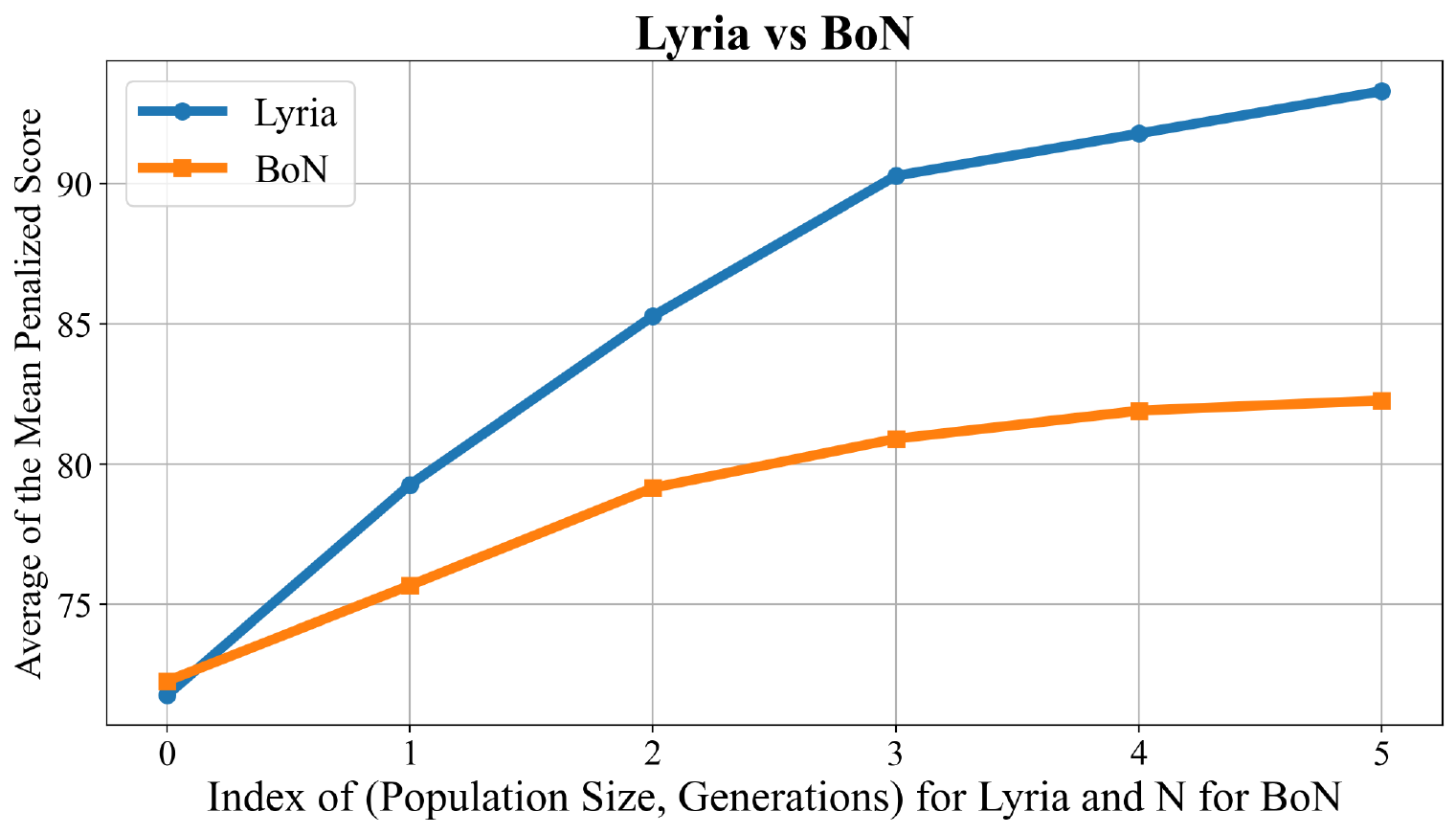}
  \caption{The figure shows the performance comparison between \LR and BoN, in which the x-axis indexes each parameter set, e.g., index $0$ means the pair of ($n_p=5$, $n_g=5$) for \LR and $N = 23$ for BoN, and the y-axis shows the corresponding score averaging across $\mathrm{SK}_{PS}$, $\mathrm{GC}_{PS}$, and $\mathrm{TSP}_{PS}$.}
  \label{fig:lyria_vs_bon}
  \end{center}
\end{figure}

This experiment investigates the impact of scaling \textit{population size} $n_p$ and \textit{generations} $n_g$ on the performance of \LR. We executed 6 experiment settings, each pairing a $n_p$ and $n_g$: \((5,5)\), \((10,10)\), \((20,20)\), \((30,30)\), \((40,40)\) and \((50,50)\). For each setting, we applied the BoN baseline for comparison, with the corresponding values of \(N\) equal to \(23\), \(80\), \(300\), \(660\), \(1160\), and \(1800\). As demonstrated in Figure~\ref{fig:lyria_vs_bon}, averaged across problems, while BoN exhibits diminishing marginal gains as parameters scaled, \LR demonstrated consistent improvements and increasingly larger performance gaps compared to BoN. We attribute the limitations of BoN to LLMs getting trapped in local optima without effective capacities to extricate themselves from it, resulting in even sampling an arbitrarily large number of answers yet still failing to yield further performance improvements. However, \LR inherently possesses the capacity to escape local optima, driving substantial performance improvements while increasing $n_p$ and $n_g$.

In addition, to disentangle the individual contribution of $n_p$ and $n_g$, we conducted 6 additional experiments settings. We fixed the $n_p$ at \(10\) while varying $n_g$ at values of \(10\), \(30\), and \(50\), and conversely fixed $n_g$ at \(10\) while adjusting $n_p$ across values of \(10\), \(30\) and \(50\). For the former, averaged across problems, the penalized scores increase by $4$, while the latter one yields $7$ gains. The modest $3$ difference between them could result from the limited diversity in smaller populations, causing offspring becoming homologous to their parents, thereby suppressing evolutionary efficacy. However, given this minor gap, we cannot exclude the possibility that it arises from stochastic variation.

\subsubsection{Oracle-Based FE VS LLM-Based FE}
\label{sec:lyria_oracl_vs_llm_fe}
This experiment seeks to explore how the performance of \LR varies when using an Oracle‑based FE versus an LLM‑based FE. We compared an Oracle-based FE with two LLM-based FEs, one built on \textit{Qwen2.5:7B-Instruct} and the other on \textit{GPT-4o-Mini}. 

We observed that, averaged across problems, the Oracle-based FE achieved a penalized score of $84$, whereas \textit{Qwen2.5:7B-Instruct} and \textit{GPT-4o-Mini} scored only $51$ and $50$, respectively. The superior result of Oracle-based FE, as expected, shows that a stronger evaluator markedly boosts the performance of \LR. The reason is obvious, that the external symbolic system can always give precise fitness to guild the evolution direction while LLMs fails to provide correct fitness resulting in evolution process oscillated. 

Additionally, it is also worth-noting that the nearly identical scores of \textit{GPT-4o-Mini} and \textit{Qwen2.5:7B-Instruct} indicate no significant difference in their evaluative capacity, although \textit{GPT-4o-Mini} demonstrates a consistent better problem solving ability than \textit{Qwen2.5:7B-instruct} as shown in Table~\ref{tab:lyria_main_results}. 


\subsubsection{Impact of ED, EP, DD}

This experiment aims to investigate the impact of the Error Detector, Experience Pool, and Deduplicator on \LR. 

For ED, we vary the \textit{maximum detected errors} $\epsilon$ at values of $0$, $3$, $6$, and $9$. For TSP, we do not observe a significant impact when increasing $\epsilon$. In contrast, for SK, as $\epsilon$ rises, the $\mathrm{SK}_{PS}$ also increased, yielding $4$ gains. For GC, increasing $\epsilon$ produced a significant $7$ improvements. We attribute the performance differences across problems to the varying efficacy of their dedicated design of ECO and EMO.

For EP, we vary the \textit{replay rate} $\rho$ at values of $0$, $0.3$, and $0.6$. We observed that varying $\rho$ did not produce significant changes in $\mathrm{GC}_{PS}$ and $\mathrm{TSP}_{PS}$. However, for SK, while setting $\rho$ to $0$ and $0.6$ yielded scores of $59$ and $62$, setting $\rho = 0.3$ produces a score of $73$, bringing up a significant improvement of $14$ and $11$ compared to the scores when $\rho=0$ and $\rho=0.6$. We attribute this discrepancy to the trade-offs of $\rho$. When $\rho$ is too low or EP is dropped, since the population of each generation evolves solely by referring to its immediate predecessors, the lack of retained historical best solutions may bias the evolutionary direction. Conversely, when $\rho$ is too high, overreliance on historical best solutions which may themselves be local optima, can homogenize the evolved population and lead to premature convergence on suboptimal solutions.

For DD, we vary the \textit{maximum deduplication attempts} $\tau$ at the values of $0$, $3$, and $6$. Averaged across problems, increasing $\tau$ does not bring up a significant improvement, which, nevertheless, is as expected and does not mean that the deduplicator is dispensable. Since the solution spaces of all the given problems are considerably immense, and when the population size remains much smaller than the solution space, it results in a low incidence of duplicate individuals, DD therefore may not be invoked. Thus, when encountering problems with comparatively smaller solution spaces, DD could effectively eliminate duplicates and thereby enhance population diversity.

\subsubsection{Impact of ECO and EMO}
\label{sec:lyria_impact_eco_emo}
This experiment aims to investigate the impact of External Crossover Operator (ECO) and External Mutation Operator (EMO) on \LR. Given the close interdependence between these two operators, rather than evaluating their efficacy in isolation, we simultaneously vary both the \textit{external crossover rate} $\xi$ and the \textit{external mutation rate} $\mu$ to investigate the efficacy of them. Thus, we construct 3 experiment settings, each pairing a $\xi$ and $\mu$: $(0,0)$, $(0.3,0.3)$, and $(0.6,0.6)$. 

For GC, raising $\xi$ and $\mu$ induces a $6.35$ performance gain by improving $\mathrm{GC}_{PS}$ from $91.07$ to $97.42$. However, for TSP, we did not observe a significant performance gain after increasing the rates. Furthermore, for SK, we observed a $6.23$ performance drop after raising rates. The results indicate that although integrating symbolic system into crossover and mutation can help improve performance, the disparity of the results illustrates that the quality of ECO and EMO designs tailored to specific problems can markedly influence performance. High-quality ECO and EMO enable \LR to evolve populations more effectively, leading to better performance. We consider that a high-quality ECO and EMO may contain, but are not limited to, extra or superior heuristics beyond what an LLM alone can provide, structural or precise constraints, or expert domain knowledge, which can not only fill the incomplete solution space that is the out-of-distribution data of LLMs but also provide appropriate crossover and mutation strategy to advance evolution. Conversely, a poor design of them may trigger performance declines. We consider that inferior operators may fail to fill the incomplete solution space or may synthesize solutions worse than their predecessors, especially when they are frequently used in the case that $\xi$ and $\mu$ are elevated, which can introduce low-quality individuals into each generation, thereby degrading performance. Therefore, a meticulous and superior design of ECO and EMO is essential for \LR.

\section{The \GF Fine-tuning Process}
\label{sec:lyria_gaft}

Building on \LR, we extend it to the fine-tuning (FT) process of LLMs and introduce the \textbf{L}yria \textbf{A}ugmented \textbf{F}ine-\textbf{T}uning, abbreviated as \textbf{\GF}, a new method that fine-tunes a weaker model to imitate the reasoning process of a stronger model that uses \LR as its reasoning framework, enabling the weaker model to reason within the \LR framework at inference time and achieve improved performance. We detail the methodology, experiment settings, and results with analysis in Section~\ref{sec:lyria_gaft_methodology},~\ref{sec:lyria_gaft_experiment_settings}, and~\ref{sec:lyria_gaft_results_analysis}, respectively. 

\subsection{Methodology}
\label{sec:lyria_gaft_methodology}

\begin{figure}[t!]
\begin{center}
  \includegraphics[width=\columnwidth]{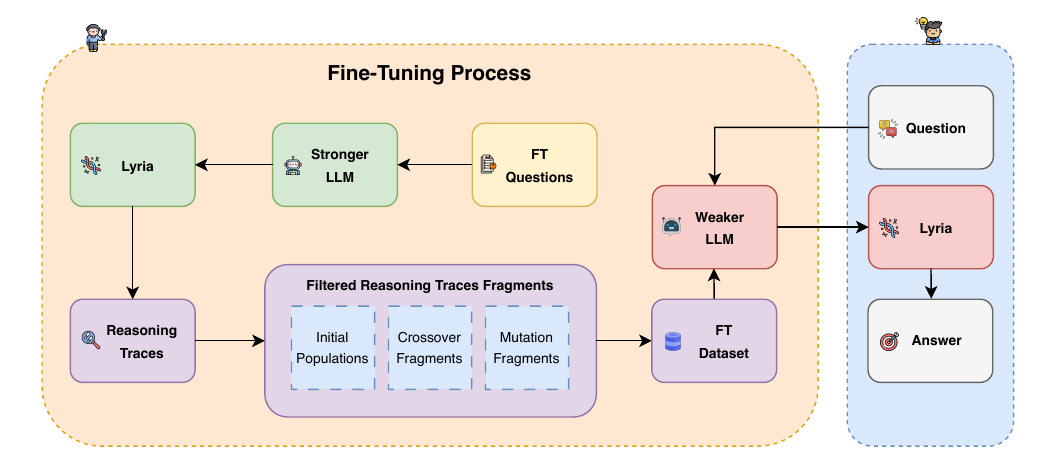}
  \caption{The overall process of \GF in which the orange dashed-border block depicts the FT process and the blue dashed-border block illustrates the inference process. The FT process begins with a set of questions generated for FT data construction, which are answered by a stronger LLM equipped with the \LR reasoning framework. During reasoning, detailed reasoning traces are collected, including the initial populations, crossover fragments, and mutation fragments. These fragments are subsequently filtered to remove suboptimal ones and retain beneficial ones, forming a FT dataset. The dataset is then used to fine-tune a weaker LLM. During inference, the fine-tuned weaker LLM employs \LR to reason over new questions and generate final answers.}
  \label{fig:lyria_gaft_ft_process}
\end{center}
\end{figure}

In \GF, as illustrated in Figure~\ref{fig:lyria_gaft_ft_process}, we initiate the process with a set of questions generated specifically for building the FT dataset. A stronger LLM then leverages \LR as its reasoning framework to reason over and generate answers to these questions. Throughout this reasoning process, we systematically record all information arising from using \LR for reasoning, referred to as reasoning traces. These traces comprise detailed information about the initial populations, as well as the information of every crossover and mutation operation, which we collectively denote as reasoning trace fragments. These information include raw responses, errors information, and fitness scores of initial populations, of both parents and child candidates in the crossover operation, and of both original and mutated candidates in the mutation operation. Subsequently, a filtering procedure is applied to filter out disturbing and inferior fragments while retaining beneficial ones. All initial populations are preserved entirely. For crossover fragments, only those in which the child outperforms both parent candidates are retained. For mutation fragments, only those where the mutated candidate surpasses its original candidate are retained. The resulting filtered fragments are then reformatted into a FT dataset, which is used to fine-tune the weaker LLM. Once fine tuning is completed, during inference, the fine-tuned weaker LLM employs \LR as its reasoning framework to reason over new questions and generate final answers.

\subsection{Experiment Settings}
\label{sec:lyria_gaft_experiment_settings}
In the experiment, we select \textit{Llama3.2:3B-Instruct}~\citep{grattafiori2024llama3herdmodels} as the weaker model, and \textit{Qwen2.5:32B-Instruct}~\citep{qwen2025qwen25technicalreport} as the stronger model. We generate 100 FT questions for each type of problems SK, GC, and TSP, respectively, prepared for building the FT dataset. The FT questions generation strictly follows the benchmark construction process described in Section~\ref{sec:lyria_benchmarks}. All FT questions are disjoint from the test sets used for evaluation. In the FT process of \GF described in Section~\ref{sec:lyria_main_experiment_settings}, for \textit{Qwen2.5:32B-Instruct} with \LR, we switch on Oracle-based FE and follow the same parameter settings as the case we set parameters of \LR described in Section~\ref{sec:lyria_methodology}. The generated FT dataset is used to fine-tune the \textit{Llama3.2:3B-Instruct} model to have a new fine-tuned model, called \textit{Llama3.2:3B-\GF}. During inference, it uses the \LR reasoning framework to answer questions. We set its parameters as that, the temperature is set to $0.7$, the maximum generated tokens to $4096$, population size to $20$, the generations to $20$, the maximum detected errors to 3, the maximum deduplication attempts to $3$, the replay rate to $0.6$, the crossover rate to $0.7$, the external crossover rate to $0.3$, the mutation rate to $0.3$, and the external mutation rate to $0.3$.

To form baselines, for each type of problems, we construct 3 new types of datasets, which are \textit{DPFT} dataset, \textit{DPFT@Full} dataset, and \textit{DPFT@Top} dataset. For the construction of \textit{DPFT}, \textit{Qwen2.5:32B-Instruct} answers each FT question once with the DP method, where the temperature is set to $0$ and max generated tokens to $4096$, following the task-specific DP templates as described in Prompt Template~\ref{pt:lyria_sudoku_dp},~\ref{pt:lyria_gc_dp}, and~\ref{pt:lyria_tsp_dp}. This yields $100$ pairs of questions and answers for each type of problems. For the construction of \textit{DPFT@Full} dataset, \textit{Qwen2.5:32B-Instruct} answers each FT question $345$ times, which is the same as the sample times $N$ of BoN method set in Section~\ref{sec:lyria_main_experiment_settings}, with the DP method, where the temperature is set to $0.7$ and the max generated tokens to $4096$, to ensure a diverse set of responses towards one question. This process produces $34500$ pairs of questions and answers for each type of problem. For the construction of the \textit{DPFT@Top} dataset, all pairs of questions and answers from the \textit{DPFT@Full} dataset are ranked according to their penalized scores and only the top $20\%$ of pairs with the highest scores are selected to form the \textit{DPFT@Top} dataset, which results in $6900$ pairs per problem type.

Based on the newly constructed datasets, for each problem type, we fine-tune \textit{Llama3.2:3B-Instruct} to obtain 3 new fine-tuned models, which are \textit{Llama3.2:3B-DPFT}, \textit{Llama3.2:3B-DPFT@Full}, and \textit{Llama3.2:3B-DPFT@Top}. For \textit{Llama3.2:3B-DPFT}, it is fine-tuned on the \textit{DPFT} dataset. For \textit{Llama3.2:3B-DPFT@Full}, it is fine-tuned on the \textit{DPFT@Full} dataset. For \textit{Llama3.2:3B-DPFT@Top}, it is fine-tuned on the \textit{DPFT@Top} dataset.

Based on the fine-tuned models, for each problem type, we form 9 baselines as follows:

\begin{enumerate}
    \item \textbf{\textit{Llama3.2:3B-Instruct} + DP}: The base model \textit{Llama3.2:3B-Instruct} is used with the DP method to answer questions. The temperature is set to $0$ and the maximum number of generated tokens to $4096$;
    
    \item \textbf{\textit{Llama3.2:3B-Instruct} + BoN}: The base model \textit{Llama3.2:3B-Instruct} is used with the BoN method to answer questions, where $N=300$, the temperature is set to $0.7$, the maximum number of generated tokens is set to $4096$, and the maximum deduplication attempts is set to 3;
    
    \item \textbf{\textit{Llama3.2:3B-Instruct} + \LR}: The base model \textit{Llama3.2:3B-Instruct} is used with the \LR method to answer questions, under the same parameter setting as in \textit{Llama3.2:3B-\GF};
    
    \item \textbf{\textit{Llama3.2:3B-DPFT} + BoN}: The fine-tuned model \textit{Llama3.2:3B-DPFT} is used with the BoN method to answer questions, under the same parameters settings as the second baseline
    
    \item \textbf{\textit{Llama3.2:3B-DPFT} + \LR}: The fine-tuned model \textit{Llama3.2:3B-DPFT} is used with the \LR method to answer questions, under the same parameters setting as the third baseline;

    \item \textbf{\textit{Llama3.2:3B-DPFT@Full} + BoN}: The fine-tuned model \textit{Llama3.2:3B-DPFT@Full} is used with the BoN method to answer questions, under the same parameters settings as the second baseline;

    \item \textbf{\textit{Llama3.2:3B-DPFT@Full} + \LR}: The fine-tuned model \textit{Llama3.2:3B-DPFT@Full} is used with the \LR method to answer questions, under the same parameters setting as the third baseline;

    \item \textbf{\textit{Llama3.2:3B-DPFT@Top} + BoN}: The fine-tuned model \textit{Llama3.2:3B-DPFT@Top} is used with the BoN method to answer questions, under the same parameters settings as the second baseline;

    \item \textbf{\textit{Llama3.2:3B-DPFT@Top} + \LR}: The fine-tuned model \textit{Llama3.2:3B-DPFT@Top} is used with the \LR method to answer questions, under the same parameters settings as the third baseline.
\end{enumerate}

\subsection{Results \& Analysis}
\label{sec:lyria_gaft_results_analysis}

We demonstrate the results for SK, GC, and TSP problems in Table~\ref{tab:lyria_gaft_sk_results},~\ref{tab:lyria_gaft_gc_results}, and~\ref{tab:lyria_gaft_tsp_results}, respectively. 

Across the results of each problem type, \textit{Llama3.2:3B-\GF} always demonstrates the best performance and consistently outperforms all 9 constructed baselines. For SK problems, it improves $\textrm{SK}_{CR}$ by $10\%$ and $\textrm{SK}_{PS}$ by $64.52$ compared to the worst approach, i.e., \textit{Llama3.2:3B-DPFT} with the BoN method, while improving $\textrm{SK}_{CR}$ by $10\%$ and $\textrm{SK}_{PS}$ by $40.30$ compared to the best approach among 9 baselines, i.e., \textit{Llama3.2:3B-DPFT@Full} with the BoN method. For GC problems, it increases $\textrm{GC}_{PS}$ by $25.02$ compared to the worst approach, i.e., \textit{Llama3.2:3B-Instruct} with the DP method, while increasing $\textrm{GC}_{PS}$ by $4.61$ compared to the two best approaches among 9 baselines, i.e., \textit{Llama3.2:3B-DPFT} with the \LR method and \textit{Llama3.2:3B-DPFT@Full} with the \LR method. For TSP problems, it enhances $\textrm{TSP}_{CR}$ by $18\%$ and $\textrm{TSP}_{PS}$ by $97.54$ compared to the worst approach, i.e., \textit{Llama3.2:3B-Instruct} with the DP method, while enhancing $\textrm{TSP}_{CR}$ by $14\%$ and $\textrm{SK}_{PS}$ by $1.51$ compared to the best approach among 9 baselines, i.e., \textit{Llama3.2:3B-DPFT@Full} with the BoN method.

\begin{table}[t!]
\centering
\begin{tabular*}{\textwidth}{@{\extracolsep{\fill}}lccc}
\toprule
\textbf{Model} & \textbf{Method} & \textbf{$\textrm{SK}_{CR}$} & \textbf{$\textrm{SK}_{PS}$} \\
\midrule
\textbf{\textit{Llama3.2:3B-\GF}} & \LR & \textbf{10} & \textbf{65.84} \\
\textit{Llama3.2:3B-Instruct} & DP & 0 & 1.83 \\
\textit{Llama3.2:3B-Instruct} & BoN & 0 & 15.04 \\
\textit{Llama3.2:3B-Instruct} & \LR & 0 & 10.45 \\
\textit{Llama3.2:3B-DPFT} & BoN & 0 & 1.32 \\
\textit{Llama3.2:3B-DPFT} & \LR & 0 & 1.38 \\
\textit{Llama3.2:3B-DPFT@Full} & BoN & 0 & 25.54 \\
\textit{Llama3.2:3B-DPFT@Full} & \LR & 0 & 12.61 \\
\textit{Llama3.2:3B-DPFT@Top} & BoN & 0 & 15.97 \\
\textit{Llama3.2:3B-DPFT@Top} & \LR & 0 & 13.45 \\
\bottomrule
\end{tabular*}
\caption{Results of our approach \GF and the constructed 9 baselines for SK.}
\label{tab:lyria_gaft_sk_results}
\end{table}

\begin{table}[t!]
\centering
\begin{tabular*}{\textwidth}{@{\extracolsep{\fill}}lccc}
\toprule
\textbf{Model} & \textbf{Method} & \textbf{$\textrm{GC}_{CR}$} & \textbf{$\textrm{GC}_{PS}$} \\
\midrule
\textbf{\textit{Llama3.2:3B-\GF}} & \LR & \textbf{0} & \textbf{92.99} \\
\textit{Llama3.2:3B-Instruct} & DP & 0 & 67.97 \\
\textit{Llama3.2:3B-Instruct} & BoN & 0 & 83.18 \\
\textit{Llama3.2:3B-Instruct} & \LR & 0 & 85.08 \\
\textit{Llama3.2:3B-DPFT} & BoN & 0 & 74.66 \\
\textit{Llama3.2:3B-DPFT} & \LR & 0 & 88.38 \\
\textit{Llama3.2:3B-DPFT@Full} & BoN & 0 & 84.44 \\
\textit{Llama3.2:3B-DPFT@Full} & \LR & 0 & 88.38 \\
\textit{Llama3.2:3B-DPFT@Top} & BoN & 0 & 84.24 \\
\textit{Llama3.2:3B-DPFT@Top} & \LR & 0 & 87.85 \\
\bottomrule
\end{tabular*}
\caption{Results of our approach \GF and the constructed 9 baselines for GC.}
\label{tab:lyria_gaft_gc_results}
\end{table}

\begin{table}[t!]
\centering
\begin{tabular*}{\textwidth}{@{\extracolsep{\fill}}lccc}
\toprule
\textbf{Model} & \textbf{Method} & \textbf{$\textrm{TSP}_{CR}$} & \textbf{$\textrm{TSP}_{PS}$} \\
\midrule
\textbf{\textit{Llama3.2:3B-\GF}} & \LR & \textbf{18} & \textbf{97.53} \\
\textit{Llama3.2:3B-Instruct} & DP & 0 & 0 \\
\textit{Llama3.2:3B-Instruct} & BoN & 0 & 10.15 \\
\textit{Llama3.2:3B-Instruct} & \LR & 0 & 31.88 \\
\textit{Llama3.2:3B-DPFT} & BoN & 0 & 90.39 \\
\textit{Llama3.2:3B-DPFT} & \LR & 0 & 91.11 \\
\textit{Llama3.2:3B-DPFT@Full} & BoN & 4 & 96.02 \\
\textit{Llama3.2:3B-DPFT@Full} & \LR & 2 & 95.11 \\
\textit{Llama3.2:3B-DPFT@Top} & BoN & 0 & 91.57 \\
\textit{Llama3.2:3B-DPFT@Top} & \LR & 0 & 91.94 \\
\bottomrule
\end{tabular*}
\caption{Results of our approach \GF and the constructed 9 baselines for TSP.}
\label{tab:lyria_gaft_tsp_results}
\end{table}

In addition, for SK problems, compared the results of \textit{Llama3.2:3B-\GF} with the results of \textit{GPT-4o-Mini}, \textit{Qwen2.5:32B-Instruct}, \textit{Mistral:7B-Instruct}, and \textit{Qwen2.5:7B-Instruct} shown in Table~\ref{tab:lyria_main_results}, the \textit{Llama3.2:3B-\GF} outperforms the best $\textrm{SK}_{CR}$ of \textit{GPT-4o-Mini}, \textit{Mistral:7B-Instruct}, and \textit{Qwen2.5:7B-Instruct} by $2\%$, $10\%$, and $10\%$ respectively, and also outperforms the best $\textrm{SK}_{PS}$ of \textit{Mistral:7B-Instruct} and \textit{Qwen2.5:7B-Instruct} by $53.84$ and $4.82$. For GC problems, the \textit{Llama3.2:3B-\GF} outperforms the best $\textrm{GC}_{PS}$ of \textit{Mistral:7B-Instruct} by $0.99$. For TSP problems, the \textit{Llama3.2:3B-\GF} outperforms the best $\textrm{TSP}_{CR}$ of \textit{GPT-4o-Mini}, \textit{Mistral:7B-Instruct}, and \textit{Qwen2.5:7B-Instruct} by $12\%$, $18\%$, and $14\%$ respectively, and also outperforms the best $\textrm{SK}_{PS}$ of \textit{GPT-4o-Mini}, \textit{Mistral:7B-Instruct}, and \textit{Qwen2.5:7B-Instruct} by $1.53$, $8.53$, and $2.53$. From the results, it indicates that, with \GF, the performance of the weaker model like \textit{Llama3.2:3B-Instruct} can even be elevated to surpass the \textit{GPT-4o-mini}. In addition, from the results, we observe that \textit{Llama3.2:3B-\GF} is the one who approaches the performance closet to the stronger model, \textit{Qwen2.5:32B-Instruct}, used to generate FT dataset, and \textit{Llama3.2:3B-\GF} even outperforms \textit{Qwen2.5:32B-Instruct} with the BoN method for $2\%$ in $\textrm{SK}_{CR}$, $5.99$ in $\textrm{GC}_{PS}$, $10\%$ in $\textrm{TSP}_{CR}$, and $0.53$ in $\textrm{TSP}_{PS}$.

Furthermore, we consider and provide several reasons to explain why \textit{Llama3.2:3B-\GF} can significantly outperform the constructed 9 baselines. Although the \textit{Llama3.2:3B-DPFT}, \textit{Llama3.2:3B-DPFT@Full}, and \textit{Llama3.2:3B-DPFT@Top} are trained on a diverse set of responses generated by \textit{Qwen2.5-32B-Intruct}, especially the latter twos, for the case that they use the BoN method, they still merely generate a diverse set of answers which may still be locked in local optima without the ability to reason beyond the local optima, and for the case they use the \LR method, since they are not fine-tuned to be able to proficiently follow and leverage the reasoning framework of \LR and they may also be disrupted by the FT process with \textit{DPFT}, \textit{DPFT@Full}, and \textit{DPFT@Top} datasets which do not include the data of \LR reasoning process, they could fail to utilize the \LR reasoning framework during the inference even explicitly equipped with \LR. However, for \textit{Llama3.2:3B-\GF}, it is fine-tuned to follow the \LR reasoning process of the stronger models, which can significantly improve its proficiency in leveraging \LR, and during inference, since \LR can help escape the local optima and explores a larger solution space, \textit{Llama3.2:3B-\GF} can then leverage \LR to reason over the questions to avoid locking in the local optima and obtain better performance.

\section{Limitations and Future Directions}
\label{sec:lyria_lim_fut}
Although our evaluation of \LR focuses on SK, GC, and TSP problems, the framework is not restricted to these domains. Importantly, this work is not to leverage LLMs to achieve SOTA performance on combinatorial optimization problems. Instead, this work aims to investigate and improve the two major issues of LLMs we concern and discuss before. With these two major issues being eased, we believe \LR can be applied in a broader range of domains. For example, in code generation domain, we can first construct the Oracle-based FE as a suite of tests and build the ECO and the EMO as abstract syntax tree-based code operations, and then use \LR to solve the code generation problems, in the way of leveraging semantic understanding of LLMs, global search ability of genetic algorithms, and complete solution space of external symbolic systems such as Oracle-based FE, ECO, and EMO. We encourage and expect the integration of \LR into these domains in future works.

In addition, in Section~\ref{sec:lyria_oracl_vs_llm_fe}, we observe a substantial performance gap between the Oracle‑based FE and the LLM‑based FE. While this unsurprisingly shows that a reliable external symbolic system can provide better guidance by giving correct fitness during evolution, it also reveals a limitation of \LR, that currently a robust Oracle-based FE is required to fully realize the effectiveness of \LR. This makes the present version of \LR unsuitable for problems where verifying solutions is difficult. Instead, it suits well for problems in which verification is easy but generating a correct solution is challenging and where an Oracle-based FE is available. For instance, the code generation tasks discussed above suits well for using \LR to solve. Nevertheless, since we do not expect, in practice, an Oracle-based FE is always available, we believe replacing the Oracle-based FE with an LLM‑based FE can greatly increase the applicability and convenience of \LR in real-world applications. Thereby, we expect future work to improve the LLM‑based FE to approach the Oracle‑based FE. 

Moreover, currently, both ECO and EMO are manually designed, and as we observe in Section~\ref{sec:lyria_impact_eco_emo}, their design could significantly contribute to the performance. For different problems or domains, to achieve effective ECOs and EMOs always requires dedicated experts to carefully craft them, often involving multiple rounds of testing and iteration. Thus, it is necessary to develop an approach that can automatically design ECOs and EMOs, thereby eliminating the burden of manual design. Such an approach would not only facilitate the rapid application and generalization of this framework across diverse domains and problems, but could also enable LLM-based agents to autonomously design ECOs and EMOs when facing different environments and tasks, and thus leverage \LR to substantially enhance their capabilities. As discussed in Section~\ref{sec:lyria_related_work}, future works may follow the approaches like LLaMEA~\citep{LLaMEA}, EoH~\citep{EoH}, and ReEvo~\citep{ReEvo} to enable automatically generating metaheuristic algorithm to construct ECOs and EMOs, which can future improve and polish the \LR reasoning framework.

Furthermore, to our best knowledge, while several recent works begin to explore approaches that integrate LLM with genetic algorithm, as mentioned in Section~\ref{sec:lyria_related_work}, these efforts have largely been focused and confined to enhancing performance within a specific domain, falling short of offering a reasoning framework that integrates of LLMs, genetic algorithms, and symbolic systems, and also a comprehensive investigation into what factors may affect and govern its effectiveness. Hence, they lack direct comparability with this work. In addition to integrating with genetic algorithms, research on hybrid frameworks that integrate LLMs with other metaheuristic algorithms remains scarce, such as integration of LLMs with particle swarm optimization or ant colony optimization. Since different metaheuristic algorithms may offer distinct advantages across various domains and problem types, further investigation along this line could yield deeper insights into the comparative benefits of integrating LLMs with different metaheuristic algorithms. However, since works that explore other LLM-driven metaheuristic frameworks in a manner similar to our approach are still lacking, we are therefore unable to make a comparison. Therefore, as no directly comparable framework currently exists, this paper concentrates more on the internal analysis to ensure that the observed performance improvements arise from the virtue of the framework itself rather than from other factors, e.g., increasing samplings, and examine the contribution and necessity of each component, while also extending the idea to fine-tuning process and proposing \GF.

Finally, while \LR exerts significant performance improvements, especially when the population size and generations are increased, it necessarily induces more LLM queries, leading to much longer response time and higher costs, i.e., $L$-times more than the DP method as discussed in Section~\ref{sec:lyria_main_experiment_baselines}. Therefore, reducing this overhead is an important goal for subsequent work.

\section{Conclusion}
\label{sec:lyria_conclusion}
In this work, we introduce \LR, a neuro-symbolic reasoning framework building on the integration of LLMs, genetic algorithms, and symbolic systems, comprising 7 essential components, to investigate and improve two major issues of LLMs, which are that LLMs trap themselves into local optima and they lack exhaustive coverage of the solution space. We conducted extensive experiments with 4 LLMs across 3 types of combinatorial reasoning problems, to show the superior effectiveness of \LR, and also conducted 7 additional ablation experiments to demonstrate how various factors affect its performance. In addition, based on \LR, we extend the ideas to the fine-tuning process of LLMs and propose \GF which enables a weaker model to imitate the reasoning process of a stronger model that operates under \LR. By conducting experiments, we demonstrate that \GF can contribute consistent and significant performance improvements against 9 constructed baselines. Furthermore, we also reveal the limitations and offer perspectives on future directions. We hope this work offers valuable insights into the integration of LLMs, genetic algorithms, and symbolic systems, and sparks further exploration in this field. 

\bibliography{colm2026_conference}
\bibliographystyle{colm2026_conference}

\clearpage
\appendix
\section{L Calculation of \LR}
\label{appendix:lyria_l_calculation}

\LR could sample an LLM up to $L$ times for a single problem. Let $n_p$ be the population size, $n_g$ be the generations, $\eta$ be the crossover rate, $\xi$ be the external crossover rate, $\kappa$ be the mutation rate, and $\mu$ be the external mutation rate. The sample times $L$ are given as:
$$
L = n_p + \bigl(n_p \cdot \eta \cdot (1-\xi) + n_p \cdot \kappa \cdot (1-\mu)\bigr) \cdot n_g
$$
Thus, for example, given $n_p = 30, n_g=15, \eta=0.7, \xi=0.3, \kappa=0.3,\mu=0.3$, we have $L = 30 + (30 \cdot 0.7 \cdot(1-0.3)+30 \cdot 0.3 \cdot (1-0.3)) \cdot 15 = 345$.

\section{Additional Results}

We demonstrate the results of all metrics for each problem type in Table~\ref{tab:lyria_additional_results_sk},~\ref{tab:lyria_additional_results_gc}, and~\ref{tab:lyria_additional_results_tsp}.

\label{appendix:lyria_additional_results}

\begin{table*}[h]
\centering
\begin{tabularx}{\textwidth}{l c @{\hspace{5em}} *{3}{X}}
\toprule
\textbf{Model} & \textbf{Method}
  & \textbf{$\mathrm{SK}_{CR}$} 
  & \textbf{$\mathrm{SK}_{SC}$}
  & \textbf{$\mathrm{SK}_{PS}$}\\
\midrule
\multirow{3}{*}{\centering GPT-4o-Mini}
& DP 
& 0 
& 43
& 39\\
& BoN  
& 6
& 74
& 73\\
& \LR
& 8
& 74 
& 73\\
\midrule
\multirow{3}{*}{\centering Qwen2.5:32B-Instruct}
& DP 
& 0 
& 35 
& 31\\
& BoN  
& 8
& 77 
& 76\\
& \LR
& 32
& 87
& 87 \\
\midrule
\multirow{3}{*}{\centering Mistral:7B-Instruct}
& DP 
& 0 
& 1
& 0 \\
& BoN  
& 0
& 11
& 5 \\
& \LR
& 0
& 16
& 12 \\
\midrule
\multirow{3}{*}{\centering Qwen2.5:7B-Instruct}
& DP 
& 0 
& 32 
& 26 \\
& BoN  
& 0
& 57
& 55 \\
& \LR
& 0
& 62 
& 61\\
\bottomrule
\end{tabularx}
\caption{The results of all metrics for SK.}
\label{tab:lyria_additional_results_sk}
\end{table*}

\begin{table*}[h]
\centering
\begin{tabularx}{\textwidth}{l c @{\hspace{3em}} *{5}{X}}
\toprule
\textbf{Model} & \textbf{Method}
  & \textbf{$\mathrm{GC}_{CR}$} 
  & \textbf{$\mathrm{GC}_{SC}$}
  & \textbf{$\mathrm{GC}_{PS}$} 
  & \textbf{$\mathrm{GC}_{CF}$}\\
\midrule
\multirow{3}{*}{\centering GPT-4o-Mini}
& DP
& 0
& 73
& 73
& 27 \\
& BoN
& 0
& 86
& 86
& 14\\
& \LR
& 0
& 97
& 97
& 4\\
\midrule
\multirow{3}{*}{\centering Qwen2.5:32B-Instruct}
& DP
& 0
& 74
& 74
& 26 \\
& BoN
& 0
& 87
& 87
& 13\\
& \LR
& 0
& 96
& 96
& 4\\
\midrule
\multirow{3}{*}{\centering Mistral:7B-Instruct}
& DP
& 0
& 100
& 0
& 0 \\
& BoN
& 0
& 86
& 84
& 15 \\
& \LR
& 0
& 93
& 92
& 7 \\
\midrule
\multirow{3}{*}{\centering Qwen2.5:7B-Instruct}
& DP
& 0
& 73
& 73
& 27 \\
& BoN
& 0
& 84
& 84
& 16 \\
& \LR
& 0
& 95
& 95
& 5\\
\bottomrule
\end{tabularx}
\caption{The results of all metrics for GC.}
\label{tab:lyria_additional_results_gc}
\end{table*}

\begin{table*}[h]
\centering
\begin{tabularx}{\textwidth}{l c @{\hspace{3.5em}} *{4}{X}}
\toprule
\textbf{Model} & \textbf{Method}
  & \textbf{$\mathrm{TSP}_{CR}$} 
  & \textbf{$\mathrm{TSP}_{PS}$} 
  & \textbf{$\mathrm{TSP}_{EDM}$}
  & \textbf{$\mathrm{TSP}_{MC}$}\\
\midrule
\multirow{3}{*}{\centering GPT-4o-Mini}
& DP
& 0
& 79
& 0.64
& 0 \\
& BoN
& 4
& 94
& 0.18
& 0\\
& \LR
& 6
& 96
& 0.13
& 0\\
\midrule
\multirow{3}{*}{\centering Qwen2.5:32B-Instruct}
& DP
& 0
& 81
& 0.58
& 0 \\
& BoN
& 8
& 97
& 0.09
& 0\\
& \LR
& 30
& 99
& 0.04
& 0\\
\midrule
\multirow{3}{*}{\centering Mistral:7B-Instruct}
& DP
& 0
& 60
& 1.20
& 2 \\
& BoN
& 0
& 80
& 0.61
& 0\\
& \LR
& 0
& 89
& 0.31
& 0 \\
\midrule
\multirow{3}{*}{\centering Qwen2.5:7B-Instruct}
& DP
& 0
& 34
& 1.97
& 5 \\
& BoN
& 0
& 88
& 0.36
& 0\\
& \LR
& 4
& 95
& 0.15
& 0\\
\bottomrule
\end{tabularx}
\caption{The results of all metrics for TSP.}
\label{tab:lyria_additional_results_tsp}
\end{table*}

\section{Prompts}
\label{appendix:lyria_prompts}

\begin{promptblock}{Sudoku LLM-Based Error Detector}
\label{pt:lyria_sudoku_ed}
===Instructions===\\
1. You are a Sudoku expert who can find the errors in a sudoku candidate solution;\\
2. Given this Sudoku puzzle and its candidate solution, you should find the errors in the candidate solution;\\
3. The correctness of the solution depends on:\\
    (1) Correct Syntax: it has a correct format, meaning each row, column, and \{subgrid\_size\}x\{subgrid\_size\} square exactly contain \{puzzle\_grid\_size\} number, and each cell is separated by space. The solution format must be in the same format as the given puzzle but there is no unfilled dot;\\
    (2) Correct Semantics: for each row, column, and \{subgrid\_size\}x\{subgrid\_size\} square, the numbers 1 to \{puzzle\_grid\_size\} should appear exactly once;\\
4. If the syntax is incorrect, the errors should be "Syntax is wrong" (Noted as Type 1 Error Msg);\\
5. If the syntax is correct, the errors should be the positions of the wrong numbers in the candidate solution with its conflict type, "row", "col", or "subgrid" (Noted as Type 2 Error Msg);\\
6. You should find all the errors in the candidate solution;\\
7. If there are no errors, the errors should be "No errors" (Noted as Type 3 Error Msg);\\
8. You can think it thoroughly in any way you want, but You MUST give the errors in the end of your thinking in the format as:\\
    (1) For Type 1 Error Msg: "Syntax is wrong" wrapped in triple backticks as a code block;\\
    (2) For Type 2 Error Msg:\\
        a. Each error is in the format as "i,j,type", where i is the row number and j is the column number starting from 0 and type is the conflict type, "row", "col", or "subgrid";\\
        b. Each error is separated by a newline;\\
        c. All errors should be wrapped in triple backticks as a code block;\\
    (3) For Type 3 Error Msg: "No errors" wrapped in triple backticks as a code block;\\
    (4) You can give comments or explanations before or after the code block but you MUST NOT give any comments or explanations in the code block;\\
===Type 1 Error Example===\\
\verb|```|\\
Syntax is wrong\\
\verb|```|\\
===Type 2 Error Example===\\
\verb|```|\\
0,0,row\\
1,0,subgrid\\
2,1,col\\
\verb|```|\\
===Type 3 Error Example===\\
\verb|```|\\
No errors\\
\verb|```|\\
===Sudoku Puzzle=== \\
\verb|```|\\
\{puzzle\}\\
\verb|```|\\
===Candidate Solution===\\
\verb|```|\\
\{candidate\}\\
\verb|```|\\
\end{promptblock}

\begin{promptblock}{Graph Coloring LLM-Based Error Detector}
\label{pt:lyria_gc_ed}
===Instructions===\\
1. You are a Graph Coloring expert who can find the errors in a graph coloring candidate solution;\\
2. Given this Graph Coloring puzzle: \\
    (1) The graph is represented by the adjacency matrix with \{n\_vertices\} vertices, in which "y" means the two vertices are adjacent and "n" means the two vertices are not adjacent;\\
    (2) The goal is to color the vertices with \{color\_count\} colors such that no two adjacent vertices have the same color;\\
3. Given its candidate solution, you should find all the errors in the candidate solution;\\
4. The correctness of the solution depends on:\\
    (1) Correct Syntax: the solution should be a list of \{n\_vertices\} integers separated by comma such as "0,1,2", each integer represents the color of the corresponding vertex, and the colors should be integers from 0 to \{color\_count - 1\};\\
    (2) Correct Semantics: for each pair of adjacent vertices, the colors of the two vertices should be different;\\
4. If the syntax is incorrect, the errors should be "Syntax is wrong" (Noted as Type 1 Error Msg);\\
5. If the syntax is correct, the errors messages should be in two types:\\
    (1) Type 2.1 Error Msg: the error msg are in the format as "i,j,color", where i is the vertex number, j is the vertex number, and color is the conflict color, and all of them are integers and separated by comma; \\
    (2) Type 2.2 Error Msg: the error msg are in a number which indicates the number of exceeded colors, such as "0" means no exceeded colors, "1" means one exceeded color, "-1" means the colors used are less than the allowed color count, and so on;\\
6. You should find all the errors in the candidate solution;\\
7. If there are no errors, the errors should be "No errors" (Noted as Type 3 Error Msg);\\
8. You can think it thoroughly in any way you want, but You MUST give the errors in the end of your thinking in the format as:\\
    (1) For Type 1 Error Msg: "Syntax is wrong" wrapped in triple backticks as a code block with the language indicator as "t1";\\
    (2) For Type 2.1 Error Msg:\\
        a. Each error is in the format as "i,j,color", where i is the vertex number, j is the vertex number, and color is the conflict color, and all of them are integers and separated by comma;\\
        b. Each error is separated by a newline;\\
        c. All errors should be wrapped in triple backticks as a code block with the language indicator as "t2.1";\\
    (3) For Type 2.2 Error Msg: the number of exceeded colors wrapped in triple backticks as a code block with the language indicator as "t2.2";\\
    (3) For Type 3 Error Msg: "No errors" wrapped in triple backticks as a code block with the language indicator as "t3";\\
    (4) You can give comments or explanations before or after the code block but you MUST NOT give any comments or explanations in the code block;\\
===Type 1 Error Msg Example===\\
\verb|```|t1\\
Syntax is wrong\\
\verb|```|\\
===Type 2.1 Error Msg Example===\\
\verb|```|t2.1\\
0,1,2\\
1,2,0\\
```\\
===Type 2.2 Error Msg Example===\\
\verb|```|t2.2\\
1\\
\verb|```|\\
===Type 3 Error Msg Example===\\
\verb|```|t3\\
No errors\\
\verb|```|\\
===Graph Adjacency Matrix===\\
\verb|```|\\
\{adjacency\_matrix\}\\
\verb|```|\\
===Candidate Solution===\\
\verb|```|\\
\{candidate\}\\
\verb|```|\\
\end{promptblock}

\begin{promptblock}{Travel Salesman Problem Error Detector}
\label{pt:lyria_tsp_ed}
===Instructions===\\
1. You are a Travel Salesman Problem expert who can find all the errors in a TSP candidate solution;\\
2. Given this Traveling Salesman Problem puzzle:\\
    (1) The distance matrix is a 2D matrix with \{n\_cities\} rows and \{n\_cities\} columns, in which each element represents the distance of traveling from the city in the row to the city in the column;\\
    (2) The goal is to find the shortest path that visits each city exactly once and returns to the origin city;\\
3. Given its candidate solution, you should find all the errors in the candidate solution;\\
4. The correctness of the solution depends on: \\
    (1) Correct Syntax: \\
        a. the solution should be a list of \{n\_cities\} integers separated by comma such as "0,1,2", each integer represents the index of the city in the path, and the indexes should be integers from 0 to \{n\_cities - 1\};\\
        b. the first and last city should be the same and should be 0, which means the path should return to the origin city which is 0;\\
        c. the index of city should be in the range from 0 to \{n\_cities - 1\};\\
    (2) Correct Semantics: \\
        a. No missing city: the path should visit each city exactly once and return to the origin city;\\
        b. Optimal path: the path should be the shortest path; \\
5. If the syntax is incorrect, the errors should be "Syntax is wrong" (Noted as Type 1 Error Msg);\\
6. If the syntax is correct, the errors messages should be in two types:\\
    (1) Type 2.1 Error Msg: the error msg are in the format of list separated by comma, where each element is a missing city in the path, such as "0,1,2", where 0, 1, 2 are the missing cities, and all of them are integers and separated by comma;\\
    (2) Type 2.2 Error Msg: the error msg are in a number which indicates the exceeded distance, such as "0" means the distance is the optimal distance, "10.5" means the distance exceeds the optimal distance by 10.5, and it should be a float;\\
6. You should find all the errors in the candidate solution;\\
7. If there are no errors, the errors should be "No errors" (Noted as Type 3 Error Msg);\\
8. You can think it thoroughly in any way you want, but You MUST give the errors in the end of your thinking in the format as:\\
    (1) For Type 1 Error Msg: "Syntax is wrong" wrapped in triple backticks as a code block with the language indicator as "t1";\\
    (2) For Type 2.1 Error Msg: \\
        a. the error msg are in the format of list separated by comma, where each element is a missing city in the path, such as "0,1,2", where 0, 1, 2 are the missing cities, and all of them are integers and separated by comma;\\
        c. the error should be wrapped in triple backticks as a code block with the language indicator as "t2.1";\\
    (3) For Type 2.2 Error Msg: the exceeded distance wrapped in triple backticks as a code block with the language indicator as "t2.2";\\
    (3) For Type 3 Error Msg: "No errors" wrapped in triple backticks as a code block with the language indicator as "t3";\\
    (4) You can give comments or explanations before or after the code block but you MUST NOT give any comments or explanations in the code block;\\
===Type 1 Error Msg Example===\\
\verb|```|t1\\
Syntax is wrong\\
\verb|```|\\
===Type 2.1 Error Msg Example===\\
\verb|```|t2.1\\
0,1,2\\
\verb|```|\\
===Type 2.2 Error Msg Example===\\
\verb|```|t2.2\\
10.5\\
\verb|```|\\
===Type 3 Error Example===\\
\verb|```|t3\\
No errors\\
\verb|```|\\
===Distance Matrix===\\
\verb|```|\\
\{distance\_matrix\}\\
\verb|```|\\
===Candidate Solution===\\
\verb|```|\\
\{candidate\}\\
\verb|```|\\
\end{promptblock}

\begin{promptblock}{Sudoku LLM-based Fitness Evaluator}
\label{pt:lyria_sudoku_llm_based_fitness_evaluator}
===Instructions===\\
1. You are a Sudoku expert who can evaluate whether a sudoku candidate solution is correct or not, or how close it is to the correct solution;\\
2. Given this Sudoku puzzle and its candidate solution, you should evaluate its score. The score is to measure how close the candidate is to the solution;\\
3. The correctness of the solution depends on: \\
    (1) Correct Syntax: it has a correct format, meaning each row, column, and \{subgrid\_size\}x\{subgrid\_size\} square exactly contain \{puzzle\_grid\_size\} number, and each cell is separated by space. The solution format must be in the same format as the given puzzle but there is no unfilled dot;\\
    (2) Correct Semantics: for each row, column, and \{subgrid\_size\}x\{subgrid\_size\} square, the numbers 1 to \{puzzle\_grid\_size\} should appear exactly once;\\
4. If the syntax is incorrect, the fitness score should be 0.0;\\
5. If the syntax is correct, the score is calculated based on the number of correct numbers in rows, columns, and subgrids, and shown in percentage. R = number of correct rows / \{puzzle\_grid\_size\} + \{delta\}, C = number of correct columns / \{puzzle\_grid\_size\} + \{delta\}, and S = number of correct subgrids / \{puzzle\_grid\_size\} + \{delta\}. The fitness score is calculated based on geometric mean as (R x C x S) ** (1/3) * 100.0, in which higher is better and 0.0 means the candidate is wrong at all while 100.0 means the candidate is correct;\\
6. In most of time, you should NOT give a score of 0.0 unless <4> are satisfied; You should give a score between 0.0 and 100.0 to indicate how close the candidate is to the correct solution;\\
7. Think it carefully and do NOT randomly guess the score;\\
8. You can think it thoroughly in any way you want, but You MUST give the score as a float number in the end of your thinking.\\
===Sudoku Puzzle===\\
\verb|```| \\
\{puzzle\}\\
\verb|```| \\
===Candidate Solution===\\
\verb|```|\\
\{candidate\}\\
\verb|```|\\
\end{promptblock}

\begin{promptblock}{Graph Coloring LLM-based Fitness Evaluator}
\label{pt:lyria_gc_llm_based_fitness_evaluator}
===Instructions===\\
1. You are a Graph Coloring expert who can evaluate whether a graph coloring candidate solution is correct or not, or how close it is to the correct solution;\\
2. Given this Graph Coloring puzzle:\\
    (1) The graph is represented by the adjacency matrix with \{n\_vertices\} vertices, in which "y" means the two vertices are adjacent and "n" means the two vertices are not adjacent;\\
    (2) The goal is to color the vertices with \{color\_count\} colors such that no two adjacent vertices have the same color; \\
3. Given its candidate solution, you should evaluate its score. The score is to measure how close the candidate is to the solution;\\
4. The correctness of the solution depends on: \\
    (1) Correct Syntax: the solution should be a list of \{n\_vertices\} integers separated by comma such as "0,1,2", each integer represents the color of the corresponding vertex, and the colors should be integers from 0 to \{color\_count - 1\};\\
    (2) Correct Semantics: for each pair of adjacent vertices, the colors of the two vertices should be different;\\
4. If the syntax is incorrect, the fitness score should be 0.0;\\
5. If the syntax is correct, the score is calculated based on:\\
    (1) Number of Conflicted Edges (noted as CE): the number of edges that two adjacent vertices have the same color;\\
    (2) Number of Exceeded Colors (noted as EC): the number of colors exceeded the allowed color count;\\
    (3) The score is now calculated as: Max(0, (1 - (CE/\{n\_edges\})) * (1 - EC/(\{n\_vertices - color\_count\}))) * 100, which means the score does not only depend on the number of conflicted edges but also the number of exceeded colors and ranges from 0.0 to 100.0;\\
6. In most of time, you should NOT give a score of 0.0 unless your are very sure; You should give a score between 0.0 and 100.0 to indicate how close the candidate is to the correct solution;\\
7. Think it carefully and do NOT randomly guess the score;\\
8. You can think it thoroughly in any way you want, but You MUST give the score as a float number in the end of your thinking.\\
===Graph Adjacency Matrix===\\
\verb|```|\\
{adjacency\_matrix\_str}\\
\verb|```|\\
===Candidate Solution===\\
\verb|```|\\
\{candidate\}\\
\verb|```|\\
\end{promptblock}

\begin{promptblock}{Travel Salesman Problem LLM-based Fitness Evaluator}
\label{pt:lyria_tsp_llm_based_fitness_evaluator}
===Instructions===\\
1. You are a Travel Salesman Problem expert who can evaluate whether a TSP candidate solution is correct or not, or how close it is to the correct solution;\\
2. Given this Traveling Salesman Problem puzzle:\\
    (1) The distance matrix is a 2D matrix with \{n\_cities\} rows and \{n\_cities\} columns, in which each element represents the distance of traveling from the city in the row to the city in the column;\\
    (2) The goal is to find the shortest path that visits each city exactly once and returns to the origin city;\\
3. Given its candidate solution, you should evaluate its score. The score is to measure how close the candidate is to the solution;\\
4. The correctness of the solution depends on: \\
    (1) Correct Syntax: \\
        a. the solution should be a list of \{n\_cities\} integers separated by comma such as "0,1,2", each integer represents the index of the city in the path, and the indexes should be integers from 0 to \{n\_cities - 1\};\\
        b. the first and last city should be the same and should be 0, which means the path should return to the origin city which is 0;\\
        c. the index of city should be in the range from 0 to \{n\_cities - 1\};\\
    (2) Correct Semantics: \\
        a. No missing city: the path should visit each city exactly once and return to the origin city;\\
        b. Optimal path: the path should be the shortest path; \\
4. If the syntax is incorrect, the fitness score should be 0.0;\\
5. If the syntax is correct, the score is calculated based on:\\
    (1) Number of Missing Cities (noted as MC): the number of missing cities in the path;\\
    (2) Used Distance (noted as UD): the total distance of the path;\\
    (3) The score is computed as follows (in range of [0...100]):\\
       1) Let base\_score = 100;\\
       2) Let OD = the sum of the shortest distances of the path; (You should try to the best to think about its optimal distance)\\
       3) Let ED = UD - OD;\\
       4) Let ED\_Multiplier = ED / OD; (calculate how much the distance exceeds the optimal distance, it MUST be in range of [0...\{DEFAULT\_EDM\}])\\
       5) distance\_excess\_ratio = ED\_Multiplier / \{DEFAULT\_EDM\}; (in range of [0...1])\\
       6) distance\_correctness = base\_score - (base\_score * distance\_excess\_ratio); (in range of [0...100])\\
       7) missing\_ratio = MC / (the length of the path - 1); (in range of [0...1])\\
       8) missing\_correctness = base\_score - (base\_score * missing\_ratio); (in range of [0...100])\\
       9) The final score is min(distance\_correctness, missing\_correctness), then clamped so it never goes below 0 or above 100.\\
6. In most of time, you should NOT give a score of 0.0 unless your are very sure; You should give a score between 0.0 and 100.0 to indicate how close the candidate is to the correct solution;\\
7. Think it carefully and do NOT randomly guess the score;\\
8. You can think it thoroughly in any way you want, but You MUST give the score as a float number in the end of your thinking.\\
===Distance Matrix===\\
\verb|```|\\
\{distance\_matrix\}\\
\verb|```|\\
===Candidate Solution===\\
\verb|```|\\
\{candidate\}\\
\verb|```|\\
\end{promptblock}

\begin{promptblock}{Sudoku LCO}
\label{pt:lyria_sudoku_lco}
===Instructions===\\
1. Given this Sudoku puzzle and these two Sudoku candidate solutions, you should thoroughly think both good and bad parts of each candidate and whether they are correct solutions to the puzzle;\\
2. If you think one of them are already correct, you can give the correct solution directly;\\
3. If you think the two candidates have good parts or bad parts, you can combine the good parts of both candidates, exclude the bad parts of both candidates, or do both of them simultaneously, aiming at creating a new candidate solution which can be better than the original candidates and approach more to the correct solution;\\
4. If you think it is not necessary to combine the two candidates, you can also give a new candidate solution which is totally different from the original candidates, aiming at approaching more to the correct solution;\\
5. After crossover, the solution should approach or become correct, which means:\\
    (1) Correct Syntax: it has a correct format, meaning each row, column, and \{subgrid\_size\}x\{subgrid\_size\} square exactly contain \{puzzle\_grid\_size\} number, and each cell is separated by space. The solution format must be in the same format as the given puzzle but there is no unfilled dot;\\
    (2) Correct Semantics: for each row, column, and \{subgrid\_size\}x\{subgrid\_size\} square, the numbers 1 to \{puzzle\_grid\_size\} should appear exactly once;\\
6. You should check the syntax carefully. If the syntax is incorrect, you should give a new solution which obey the rule "correct syntax";\\
7. You should check the semantics carefully. If the semantics is incorrect, you should give a new solution which obey the rule "correct semantics";\\
8. You can think whatever way you want, but at the end of thinking, the final solution should be given and written in the same format as the puzzle wrapped in triple backticks as a code block;\\
9. You can give thinking steps or explanation before or after code block but you MUST NOT give any comments or explanations in the code block;\\
===Sudoku Puzzle===\\
\verb|```|\\
\{puzzle\}\\
\verb|```|\\
===Candidate Solution 1===\\
\verb|```|\\
\{c1\}\\
\verb|```|\\
Score of Candidate Solution 1: \{s1\} (0.0 means the candidate is wrong at all while 100.0 means the candidate is correct)\\
Errors of Candidate Solution 1: \\
\{c1\_error\}\\
===Candidate Solution 2===\\
\verb|```|\\
\{c2\}\\
\verb|```|\\
Score of Candidate Solution 2: \{s2\} (0.0 means the candidate is wrong at all while 100.0 means the candidate is correct)\\
Errors of Candidate Solution 2: \\
\{c2\_error\}\\
~\\
Now, keep the scores and errors in mind and think about how to combine the two candidates to create a new candidate solution that is better than the original candidates.\\
You can think in any way but you must finally give a candidate solution wrapped in triple backticks as a code block in the same format as the puzzle:\\
\end{promptblock}

\begin{promptblock}{Graph Coloring LCO}
\label{pt:lyria_gc_lco}
===Instructions===\\
1. Given this Graph Coloring puzzle:\\
    (1) The graph is represented by the adjacency matrix with \{n\_vertices\} vertices, in which "y" means the two vertices are adjacent and "n" means the two vertices are not adjacent;\\
    (2) The goal is to color the vertices with \{color\_count\} colors such that no two adjacent vertices have the same color; \\
2. Given these two candidate solutions, you should thoroughly think both good and bad parts of each candidate and whether they are correct solutions to the puzzle;\\
3. If you think one of them are already correct, you can give the correct solution directly;\\
4. If you think the two candidates have good parts or bad parts, you can combine the good parts of both candidates, exclude the bad parts of both candidates, or do both of them simultaneously, aiming at creating a new candidate solution which can be better than the original candidates and approach more to the correct solution;\\
5. If you think it is not necessary to combine the two candidates, you can also give a new candidate solution which is totally different from the original candidates, aiming at approaching more to the correct solution;\\
6. After crossover, the solution should approach or become correct, which means:\\
    (1) Correct Syntax: the solution should be a list of \{n\_vertices\} integers separated by comma such as "0,1,2", each integer represents the color of the corresponding vertex, and the colors should be integers from 0 to \{color\_count - 1\};\\
    (2) Correct Semantics: for each pair of adjacent vertices, the colors of the two vertices should be different;\\
7. You should check the syntax carefully. If the syntax is incorrect, you should give a new solution which obey the rule "correct syntax";\\
8. You should check the semantics carefully. If the semantics is incorrect, you should give a new solution which obey the rule "correct semantics";\\
9. You can think whatever way you want, but at the end of thinking, the final solution should be given and written in a list of integers separated by comma wrapped in triple backticks as a code block;\\
10. You can give thinking steps or explanation before or after code block but you MUST NOT give any comments or explanations in the code block;\\
===Graph Adjacency Matrix===\\
\verb|```|\\
\{adjacency\_matrix\}\\
\verb|```|\\
===Candidate Solution 1===\\
\verb|```|\\
\{c1\}\\
\verb|```|\\
Score of Candidate Solution 1: \{s1\} (0.0 means the candidate is wrong at all while 100.0 means the candidate is correct)\\
Errors of Candidate Solution 1: \\
\{c1\_error\}\\
===Candidate Solution 2===\\
\verb|```|\\
\{c2\}\\
\verb|```|\\
Score of Candidate Solution 2: \{s2\} (0.0 means the candidate is wrong at all while 100.0 means the candidate is correct)\\
Errors of Candidate Solution 2: \\
\{c2\_error\}\\
~\\
Now, keep the scores and errors in mind and think about how to combine the two candidates to create a new candidate solution that is better than the original candidates.\\
You can think in any way but you must finally give a candidate solution as a list of integers separated by comma wrapped in triple backticks as a code block:\\
\end{promptblock}

\begin{promptblock}{Travel Salesman Problem LCO}
\label{pt:lyria_tsp_lco}
===Instructions===\\
1. Given this Traveling Salesman Problem puzzle:\\
    (1) The distance matrix is a 2D matrix with \{n\_cities\} rows and \{n\_cities\} columns, in which each element represents the distance of traveling from the city in the row to the city in the column;\\
    (2) The goal is to find the shortest path that visits each city exactly once and returns to the origin city;\\
2. Given these two candidate solutions, you should thoroughly think both good and bad parts of each candidate and whether they are correct solutions to the puzzle;\\
3. If you think one of them are already correct, you can give the correct solution directly;\\
4. If you think the two candidates have good parts or bad parts, you can combine the good parts of both candidates, exclude the bad parts of both candidates, or do both of them simultaneously, aiming at creating a new candidate solution which can be better than the original candidates and approach more to the correct solution;\\
5. If you think it is not necessary to combine the two candidates, you can also give a new candidate solution which is totally different from the original candidates, aiming at approaching more to the correct solution;\\
6. After crossover, the solution should approach or become correct, which means:\\
    (1) Correct Syntax: \\
        a. the solution should be a list of \{n\_cities\} integers separated by comma such as "0,1,2", each integer represents the index of the city in the path, and the indexes should be integers from 0 to \{n\_cities - 1\};\\
        b. the first and last city should be the same and should be 0, which means the path should return to the origin city which is 0;\\
        c. the index of city should be in the range from 0 to \{n\_cities - 1\};\\
    (2) Correct Semantics: \\
        a. No missing city: the path should visit each city exactly once and return to the origin city;\\
        b. Optimal path: the path should be the shortest path; \\
7. You should check the syntax carefully. If the syntax is incorrect, you should give a new solution which obey the rule "correct syntax";\\
8. You should check the semantics carefully. If the semantics is incorrect, you should give a new solution which obey the rule "correct semantics";\\
9. You can think whatever way you want, but at the end of thinking, the final solution should be given and written in a list of integers separated by comma wrapped in triple backticks as a code block;\\
10. You can give thinking steps or explanation before or after code block but you MUST NOT give any comments or explanations in the code block;\\
===Distance Matrix===\\
\verb|```|\\
\{distance\_matrix\}\\
\verb|```|\\
===Candidate Solution 1===\\
\verb|```|\\
\{c1\}\\
\verb|```|\\
Score of Candidate Solution 1: \{s1\} (0.0 means the candidate is wrong at all while 100.0 means the candidate is correct)\\
Errors of Candidate Solution 1: \\
\{c1\_error\}\\
===Candidate Solution 2===\\
\verb|```|\\
\{c2\}\\
\verb|```|\\
Score of Candidate Solution 2: \{s2\} (0.0 means the candidate is wrong at all while 100.0 means the candidate is correct)\\
Errors of Candidate Solution 2: \\
\{c2\_error\}\\
~\\
Now, keep the scores and errors in mind and think about how to combine the two candidates to create a new candidate solution that is better than the original candidates.\\
You can think in any way but you must finally give a candidate solution as a list of integers separated by comma wrapped in triple backticks as a code block:\\
\end{promptblock}

\begin{promptblock}{Sudoku LMO}
\label{pt:lyria_sudoku_lmo}
===Instructions===\\
1. Given this Sudoku puzzle and this Sudoku candidate solution, you should thoroughly think about the good and bad parts of the candidate and whether it is a correct solution to the puzzle;\\
2. If you think the candidate is already correct, you can give the correct solution directly;\\
3. If you think the candidate has bad parts, you can change or improve the bad parts to make it good, aiming at creating a new candidate solution which can be better than the original candidate and approach more to the correct solution;\\
4. If you think it is not necessary to change the candidate, you can also give a new candidate solution which is totally different from the original candidate, aiming at approaching more to the correct solution;\\
5. After mutation, the solution should approach or become correct, which means:\\
    (1) Correct Syntax: it has a correct format, meaning each row, column, and \{subgrid\_size\}x\{subgrid\_size\} square exactly contain \{puzzle\_grid\_size\} number, and each cell is separated by space. The solution format must be in the same format as the given puzzle but there is no unfilled dot;\\
    (2) Correct Semantics: for each row, column, and \{subgrid\_size\}x\{subgrid\_size\} square, the numbers 1 to \{puzzle\_grid\_size\} should appear exactly once;\\
6. You should check the syntax carefully. If the syntax is incorrect, you should give a new solution which obey the rule "correct syntax";\\
7. You should check the semantics carefully. If the semantics is incorrect, you should give a new solution which obey the rule "correct semantics";\\
8. You can think whatever way you want, but at the end of thinking, the final solution should be given and written in the same format as the puzzle wrapped in triple backticks as a code block;\\
9. You can give thinking steps or explanation before or after code block but you MUST NOT give any comments or explanations in the code block;\\
===Sudoku Puzzle===\\
\verb|```|\\
\{puzzle\}\\
\verb|```|\\
===Candidate Solution===\\
\verb|```|\\
\{candidate\}\\
\verb|```|\\
Score of Candidate Solution: \{score\} (0.0 means the candidate is wrong at all while 100.0 means the candidate is correct)\\
Errors of Candidate Solution: \\
\{error\}\\
~\\
Now, keep the score and errors in mind and think about how to change the candidate to create a new candidate solution that is better than the original candidate.\\
You can think in any way but you must finally give a candidate solution wrapped in triple backticks as a code block in the same format as the puzzle: \\
\end{promptblock}

\begin{promptblock}{Graph Coloring LMO}
\label{pt:lyria_gc_lmo}
===Instructions===\\
1. Given this Graph Coloring puzzle:\\
    (1) The graph is represented by the adjacency matrix with \{n\_vertices\} vertices, in which "y" means the two vertices are adjacent and "n" means the two vertices are not adjacent;\\
    (2) The goal is to color the vertices with \{color\_count\} colors such that no two adjacent vertices have the same color; \\
2. Given this candidate solution, you should thoroughly think about the good and bad parts of the candidate and whether it is a correct solution to the puzzle;\\
3. If you think the candidate is already correct, you can give the correct solution directly;\\
4. If you think the candidate has bad parts, you can change or improve the bad parts to make it good, aiming at creating a new candidate solution which can be better than the original candidate and approach more to the correct solution;\\
5. If you think it is not necessary to change the candidate, you can also give a new candidate solution which is totally different from the original candidate, aiming at approaching more to the correct solution;\\
6. After mutation, the solution should approach or become correct, which means:\\
    (1) Correct Syntax: the solution should be a list of \{n\_vertices\} integers separated by comma such as "0,1,2", each integer represents the color of the corresponding vertex, and the colors should be integers from 0 to \{color\_count - 1\};\\
    (2) Correct Semantics: for each pair of adjacent vertices, the colors of the two vertices should be different;\\
7. You should check the syntax carefully. If the syntax is incorrect, you should give a new solution which obey the rule "correct syntax";\\
8. You should check the semantics carefully. If the semantics is incorrect, you should give a new solution which obey the rule "correct semantics";\\
9. You can think whatever way you want, but at the end of thinking, the final solution should be given and written in a list of integers separated by comma wrapped in triple backticks as a code block;\\
10. You can give thinking steps or explanation before or after code block but you MUST NOT give any comments or explanations in the code block;\\
===Graph Adjacency Matrix===\\
\verb|```|\\
\{adjacency\_matrix\}\\
\verb|```|\\
===Candidate Solution===\\
\verb|```|\\
\{candidate\}\\
\verb|```|\\
Score of Candidate Solution: \{score\} (0.0 means the candidate is wrong at all while 100.0 means the candidate is correct)\\
Errors of Candidate Solution: \\
\{error\}\\
~\\
Now, keep the score and errors in mind and think about how to change the candidate to create a new candidate solution that is better than the original candidate.\\
You can think in any way but you must finally give a candidate solution as a list of integers separated by comma wrapped in triple backticks as a code block: \\
\end{promptblock}

\begin{promptblock}{Travel Salesman Problem LMO}
\label{pt:lyria_tsp_lmo}
===Instructions===\\
1. Given this Traveling Salesman Problem puzzle:\\
    (1) The distance matrix is a 2D matrix with \{n\_cities\} rows and \{n\_cities\} columns, in which each element represents the distance of traveling from the city in the row to the city in the column;\\
    (2) The goal is to find the shortest path that visits each city exactly once and returns to the origin city;\\
2. Given this candidate solution, you should thoroughly think about the good and bad parts of the candidate and whether it is a correct solution to the puzzle;\\
3. If you think the candidate is already correct, you can give the correct solution directly;\\
4. If you think the candidate has bad parts, you can change or improve the bad parts to make it good, aiming at creating a new candidate solution which can be better than the original candidate and approach more to the correct solution;\\
5. If you think it is not necessary to change the candidate, you can also give a new candidate solution which is totally different from the original candidate, aiming at approaching more to the correct solution;\\
6. After mutation, the solution should approach or become correct, which means:\\
    (1) Correct Syntax: \\
        a. the solution should be a list of \{n\_cities\} integers separated by comma such as "0,1,2", each integer represents the index of the city in the path, and the indexes should be integers from 0 to \{n\_cities - 1\};\\
        b. the first and last city should be the same and should be 0, which means the path should return to the origin city which is 0;\\
        c. the index of city should be in the range from 0 to \{n\_cities - 1\};\\
    (2) Correct Semantics: \\
        a. No missing city: the path should visit each city exactly once and return to the origin city;\\
        b. Optimal path: the path should be the shortest path; \\
7. You should check the syntax carefully. If the syntax is incorrect, you should give a new solution which obey the rule "correct syntax";\\
8. You should check the semantics carefully. If the semantics is incorrect, you should give a new solution which obey the rule "correct semantics";\\
9. You can think whatever way you want, but at the end of thinking, the final solution should be given and written in a list of integers separated by comma wrapped in triple backticks as a code block;\\
10. You can give thinking steps or explanation before or after code block but you MUST NOT give any comments or explanations in the code block;\\
===Distance Matrix===\\
\verb|```|\\
\{distance\_matrix\}\\
\verb|```|\\
===Candidate Solution===\\
\verb|```|\\
\{candidate\}\\
\verb|```|\\
Score of Candidate Solution: \{score\} (0.0 means the candidate is wrong at all while 100.0 means the candidate is correct)\\
Errors of Candidate Solution: \\
\{error\}\\
~\\
Now, keep the score and errors in mind and think about how to change the candidate to create a new candidate solution that is better than the original candidate.\\
You can think in any way but you must finally give a candidate solution as a list of integers separated by comma wrapped in triple backticks as a code block: \\
\end{promptblock}

\begin{promptblock}{Sudoku Direct Prompting}
\label{pt:lyria_sudoku_dp}
===Instructions===\\
1. Given this Sudoku puzzle, you should fill in the missing numbers represented by dots;\\
2. The solution should be correct, which means:\\
    (1) Correct Syntax: it has a correct format, meaning each row, column, and \{subgrid\_size\}x\{subgrid\_size\} square exactly contain \{puzzle\_grid\_size\} number, and each cell is separated by space. The solution format must be in the same format as the given puzzle but there is no unfilled dot;\\
    (2) Correct Semantics: for each row, column, and \{subgrid\_size\}x\{subgrid\_size\} square, the numbers 1 to \{puzzle\_grid\_size\} should appear exactly once;\\
3. The puzzle is guaranteed to have a unique solution;\\
4. You should check the syntax carefully. If the syntax is incorrect, you should give a new solution which obey the rule "correct syntax";\\
5. You should check the semantics carefully. If the semantics is incorrect, you should give a new solution which obey the rule "correct semantics";\\
6. You can think whatever way you want, but at the end of thinking, the final solution should be given and written in the same format as the puzzle wrapped in triple backticks as a code block;\\
7. You can give thinking steps or explanation before or after code block but you MUST NOT give any comments or explanations in the code block;\\
===Sudoku Puzzle===\\
\verb|```|\\
\{puzzle\}\\
\verb|```| \\
\end{promptblock}

\begin{promptblock}{Graph Coloring Direct Prompting}
\label{pt:lyria_gc_dp}
===Instructions===\\
1. Given this Graph Coloring puzzle:\\
    (1) The graph is represented by the adjacency matrix with \{n\_vertices\} vertices, in which "y" means the two vertices are adjacent and "n" means the two vertices are not adjacent;\\
    (2) The goal is to color the vertices with \{color\_count\} colors such that no two adjacent vertices have the same color; \\
2. The solution should be correct, which means:\\
    (1) Correct Syntax: the solution should be a list of \{n\_vertices\} integers separated by comma such as "0,1,2", each integer represents the color of the corresponding vertex, and the colors should be integers from 0 to \{color\_count - 1\};\\
    (2) Correct Semantics: for each pair of adjacent vertices, the colors of the two vertices should be different;\\
3. You should check the syntax carefully. If the syntax is incorrect, you should give a new solution which obey the rule "correct syntax";\\
4. You should check the semantics carefully. If the semantics is incorrect, you should give a new solution which obey the rule "correct semantics";\\
5. You can think whatever way you want, but at the end of thinking, the final solution should be given and written in a list of integers separated by comma wrapped in triple backticks as a code block;\\
6. You can give thinking steps or explanation before or after code block but you MUST NOT give any comments or explanations in the code block;\\
===Graph Adjacency Matrix===\\
\verb|```|\\
\{adjacency\_matrix\}\\
\verb|```|\\
\end{promptblock}

\begin{promptblock}{Travel Salesman Problem Direct Prompting}
\label{pt:lyria_tsp_dp}
===Instructions===\\
1. Given this Traveling Salesman Problem puzzle:\\
    (1) The distance matrix is a 2D matrix with \{n\_cities\} rows and \{n\_cities\} columns, in which each element represents the distance of traveling from the city in the row to the city in the column;\\
    (2) The goal is to find the shortest path that visits each city exactly once and returns to the origin city;\\
2. The solution should be correct, which means:\\
    (1) Correct Syntax: \\
        a. the solution should be a list of \{n\_cities\} integers separated by comma such as "0,1,2", each integer represents the index of the city in the path, and the indexes should be integers from 0 to \{n\_cities - 1\};\\
        b. the first and last city should be the same and should be 0, which means the path should return to the origin city which is 0;\\
        c. the index of city should be in the range from 0 to \{n\_cities - 1\};\\
    (2) Correct Semantics: \\
        a. No missing city: the path should visit each city exactly once and return to the origin city;\\
        b. Optimal path: the path should be the shortest path; \\
3. You should check the syntax carefully. If the syntax is incorrect, you should give a new solution which obey the rule "correct syntax";\\
4. You should check the semantics carefully. If the semantics is incorrect, you should give a new solution which obey the rule "correct semantics";\\
5. You can think whatever way you want, but at the end of thinking, the final solution should be given and written in a list of integers separated by comma wrapped in triple backticks as a code block;\\
6. You can give thinking steps or explanation before or after code block but you MUST NOT give any comments or explanations in the code block;\\
===Distance Matrix===\\
\verb|```|\\
\{distance\_matrix\}\\
\verb|```|\\
\end{promptblock}
\end{document}